\def\year{2022}\relax
\documentclass[letterpaper]{article} 
\usepackage{aaai22}  
\usepackage{times}  
\usepackage{helvet}  
\usepackage{courier}  
\usepackage[hyphens]{url}  
\usepackage{graphicx} 
\usepackage{natbib}  
\usepackage{caption} 
\DeclareCaptionStyle{ruled}{labelfont=normalfont,labelsep=colon,strut=off} 
\usepackage{algorithm}
\usepackage{algorithmic}
\usepackage[utf8]{inputenc} 
\usepackage[T1]{fontenc}    
\usepackage{hyperref}       
\usepackage{url}            
\usepackage{booktabs}       
\usepackage{amsfonts}       
\usepackage{nicefrac}       
\usepackage{microtype}      
\usepackage{xcolor}         
\usepackage{times}
\usepackage{epsfig}
\usepackage{graphicx,epstopdf}
\usepackage{amsmath}
\usepackage{breqn}
\usepackage{amssymb, amsmath,amsfonts,amsthm,amssymb,mathtools,graphicx,bm}
\usepackage{subfigure}
\newcommand{\bx}{\mathbf{x}}

\newcommand{\by}{\mathbf{y}}

\newcommand{\cR}{\mathcal{R}}
\newcommand{\bh}{\mathbf{h}}

\usepackage{comment}
\usepackage{todonotes}
\usepackage{wrapfig}
\usepackage{lipsum}
\usepackage{wrapfig}

\usepackage{newfloat}
\usepackage{listings}
\floatstyle{ruled}
\newfloat{listing}{tb}{lst}{}
\floatname{listing}{Listing}
\title{Latent Time Neural Ordinary Differential Equations}

\author{
    Srinivas Anumasa,
    P.K. Srijith
}
\affiliations{
    \textsuperscript{}Computer Science and Engineering\\
    Indian Institute of Technology Hyderabad, India\\
    cs16resch11004@iith.ac.in,srijith@cse.iith.ac.in\\
}

\begin{document}
\maketitle

\begin{abstract}
Neural ordinary differential equations (NODE) have been proposed as a continuous depth generalization  to popular deep learning models such as Residual networks (ResNets). They provide  parameter efficiency and automate the model selection process in deep learning models to some extent. However, they lack the much-required uncertainty modelling and robustness capabilities which are crucial for their use in several real-world applications such as autonomous driving and healthcare. We propose a novel and unique approach to model uncertainty in NODE by considering a distribution over the end-time $T$ of the ODE solver. The proposed approach, latent time NODE (LT-NODE), treats $T$ as  a latent variable and apply Bayesian learning to obtain a posterior distribution over $T$  from the data. In particular, we use variational inference to learn an approximate posterior and the model parameters.  Prediction is done by considering the NODE representations from different samples of the posterior and can be done efficiently using a single forward pass. As $T$ implicitly defines the depth of  a NODE,   posterior distribution over $T$ would also help in model selection in NODE. We also propose,  adaptive latent time NODE (ALT-NODE), which allow each data point to have a distinct posterior distribution over end-times.  ALT-NODE uses amortized variational inference to learn an  approximate posterior using inference networks.  We demonstrate the effectiveness of the proposed approaches in modelling uncertainty and robustness through experiments on  synthetic  and several real-world image classification data.
\end{abstract}
\section{Introduction}
Deep learning models such as Residual networks (ResNet)~\citep{he2016resnet}  
have brought advances in several  computer vision tasks~\citep{ren17,he20,wang2019hierarchical}. They used  skip connections to allow the models to grow deeper and improve performance  without suffering from the  vanishing gradient problem. 
Recently, neural ordinary differential equations (NODEs)~\citep{node} were proposed as a continuous depth generalization  to ResNets. The  feature computations in ResNet can be seen as  solving an ordinary differential equation (ODE) with Euler method~\citep{lu2018beyond,haber2017stable,ruthotto2019partial}. Here, the ODE is  parameterized  by a neural network and the NODE can grow to an arbitrary depth as defined by the end-time $T$.  
It was shown that NODE is more robust~\citep{robustode} than traditional deep learning models, and is invertible, parameter efficient and maintains a constant  memory cost with respect to growth in depth.  

Modeling uncertainty is paramount for many high-risk applications such healthcare~\citep{ker2017deepmedical} and autonomous driving vehicles~\citep{fridman2019autonomous}.  However, standard NODE models compute a point estimate of predictions which fail to  capture uncertainty in  predictions.  Like ResNets, they tend to make high confidence wrong predictions on  out-of-sample observations~\citep{anumasa2021improving}, restricting their use in  high-risk  applications. There exist very few works trying to address the uncertainty in NODE~\citep{anumasa2021improving,kong2020sdenet,dandekar2021bayesian} and their uncertainty modelling capabilities are restricted by   architectural and training assumptions.  Though, NODE models have addressed the model selection in deep learning to a great extent, it still require the user to define the parameters such as end time $T$ to the  ODE solver.   This implicitly determines the depth of the NODE.  In this work, we propose a unique approach to model uncertainty in  NODE,  latent time neural ODE (LT-NODE),  which addresses these drawbacks by learning a distribution over end-time $T$. 

LT-NODE is based on the idea of capturing uncertainty by treating the  end time $T$ as a latent variable. This allows us to define a distribution over $T$ and the representations of the data point at different values of $T$ sampled from the distribution provides an estimate of uncertainty. To  capture uncertainty and to obtain a good generalization capability,  it is important to learn the distribution over $T$ from the data and we employ Bayesian inference techniques such as variational inference to learn an approximate posterior.  Consequently,  the posterior over $T$  will also help in  addressing the model selection in NODE in determining an appropriate end time.  The proposed approach can get uncertainty estimates using single forward pass and is very efficient unlike other uncertainty modelling techniques which require  multiple model evaluations. Moreover, it provides uncertainty estimates with hardly any increase in the number of parameters, as it only needs to estimate two additional parameters associated  with the variational posterior. 

Recently, it was shown that a NODE with a different depth (end-time) for different data points can overcome  the drawbacks of standard NODEs, for e.g., in solving \textit{ concentric annuli} and \textit{reflection} tasks~\citep{augmentedode,massaroli2020dissecting}. Inspired by this, we propose a variant, adaptive latent time neural ODE (ALT-NODE), which allows each sample to have a separate posterior distribution over end-time. To learn the posterior, we consider an amortized variational inference, where we specify an inference network which provides the variational approximation over $T$ for each sample. 
We develop ALT-NODEs which also do predictions efficiently, by requiring only one forward pass through the model. Moreover, the proposed uncertainty estimation techniques for neural ODEs are generic and can be   applied to several recent variants of the NODE model and architectures.  
We demonstrate the superior uncertainty modelling capability of LT-NODE and ALT-NODE under different experimental setups on synthetic  and several real-world image classification data sets  such as CIFAR10, SVHN, MNIST, and  F-MNIST.  
Our main contributions can be summarized as follows.
\begin{enumerate}
    \item We propose a novel and unique approach to  model uncertainty in NODE by treating end-time $T$ as latent and learns a posterior distribution over end-times  which also aids in model selection. 
    \item We  propose a variant which learns input dependent posterior distribution  over latent end-times. 
    \item We develop variational inference and amortized variational inference techniques for the proposed model to learn an approximate posterior distribution over latent end-times. 
    \item We demonstrate the uncertainty and robustness modelling capability of the proposed models on different experimental setups and on several image classification data sets.
\end{enumerate}

\section{Related work}
Neural ODEs~\citep{node} are continuous depth generalization of ResNets~\citep{he2016resnet} and was shown to provide competitive results on several image classification tasks. Recently, several NODE variants were proposed which improved the generalization performance in NODE. For instance, \citet{ACA} addressed the flaws in  the adjoint sensitive method used  to learn parameters in NODE to improve gradient computation and performance. Augmented NODE (ANODE)~\cite{anode} augmented the latent layers with additional dimension and was found to be more effective than NODE  in solving complex problems such as \textit{concentric annuli} and \textit{reflection}.  \citep{massaroli2020dissecting} addressed it by assuming depth of the NODE to be adaptive and data dependent. They also provide NODE variants which generalizes ANODE to consider data dependent and higher order augmentation. There are NODE variants which improves performance by letting  the parameters to change over time~\citep{massaroli2020dissecting,zhang19} or through  regularization~\citep{pmlr20,ghosh20}. However, very few works aim to address the lack of uncertainty modelling and robustness capabilities in the neural ODE models. Although NODE~\citep{robustode} was shown to be more robust than  similar ResNet architecture,  they lack the required robustness and  uncertainty modelling capabilities~\citep{anumasa2021improving}.

NODE-GP  replaced the fully connected neural  network layer in NODE with Gaussian processes to improve uncertainty and robustness  capabilities in NODE~\citep{anumasa2021improving}. SDE-Net~\citep{kong2020sdenet} tries to address this by using the framework of  stochastic differential equations. SDE-Net uses  an additional  diffusion network which learns to provide a high diffusion for the computed state trajectories of the data outside the training distribution. However, SDE-Net suffers some drawbacks in  that it requires an additional diffusion network which needs to be trained explicitly on an  out-of-distribution (OOD) data and require multiple forward passes through the model to get uncertainty estimates. Explicit training on an OOD data is practically infeasible for several applications. Concurrent to our work, Bayesian neural ODE~\citep{dandekar2021bayesian} proposes to model uncertainty using the standard technique of  learning a distribution  over weights through the black-box inference techniques based on Markov chain Monte Carlo (MCMC) methods. We propose an uncertainty modelling technique unique to NODE and yet generalizable to several NODE architectures, where the uncertainty is modeled by considering a  distribution over end-times. The proposed approach requires only a single forward pass through the model to obtain uncertainty aware predictive probability and  models the uncertainty with just 2 additional parameters (variational parameters). This fully Bayesian approach which computes posterior over end-time is different from \citet{ghosh20} which uses random end-times only as a regularization technique during training and does not model uncertainty over predictions. Our approach to model uncertainty corroborates well with some recent advances in modelling uncertainty in discrete depth networks~\citep{dikov19,antoran2020depth,wenzel20}. They show that uncertainty can be modelled effectively  by considering representations from different layers or through distributions over hyper-parameters or architectures of a deep neural network. The proposed approach differs from them as  NODE requires different probabilistic modelling, learning objective and training due to its continuous depth character. In addition, the use of amortized variational inference to learn input specific posterior distribution over end-times, further makes the proposed approach   unique and novel.  

The proposed approaches are different from the latent NODEs~\cite{rubanova2019latent,ode2vae} which are generative NODEs used for modeling the latent state dynamics associated with time series data. They assume  the initial  state to be latent and a  posterior distribution learnt over the initial state is used to generate the time series data. In contrast, we address the regression and classification problems with a fixed initial state (input image or a transformation of it).  Hence, we consider a distribution over latent end-times and associated representations in a  feed forward NODE to model uncertainty. However, this may not be a suitable for time series data where measurement times are observed.

\section{Background}
We consider a supervised learning problem and let $\mathcal{D} =  \{X,\by\} = \{(\bx_i,y_i)\}_{i=1}^{N}$ be the set of training data points with input $\bx_i \in \cR^D$ and $y_i \in \{1,\ldots,C\}$  for a classification and $y_i \in \cR$ for regression. 
For a discrete deep learning model, the hidden representation at layer $l$ is denoted as $\bh_{l}$. We consider the hidden layer representations obtained through neural ODE transformations for a point $\bx$ at time $t$ as $\bh_{\bx}(t)$, where $\bh_{\bx}(t) \in \cR^H$.   The neural network transformations defining the ODE (NODE block)  is denoted as $f(\bh_{\bx}(t),t, \pmb{\theta}_h)$, with $\pmb{\theta}_h$ being the neural network parameters. Typically, a NODE block is a stack of convolution layers or fully connected layers with nonlinear activation functions. The fully connected neural network (FCNN) transforming the hidden representation to a probability over the output $y$  is represented as $g_y(\bh_{\bx}(t), \pmb{\theta}_g)$, with parameters $\pmb{\theta}_g$. We denote $\pmb{\theta}$ to represent all the parameters in the NODE including the initial down-sampling block. 

\subsection{Neural Ordinary Differential Equations}
ResNets transform the input to an output using a sequence of neural network transformations $f(\cdot)$ with skip connections between layers. 
The operations  on a hidden representation  $\mathbf{h}_t$ to obtain $\mathbf{h}_{t+1}$ in ResNets can be  expressed as $\bh_{t+1} = \bh_t + f(\bh_t, \pmb{\theta}_h)$. 
Neural ordinary differential equations (NODEs) show that a sequence of such transformations can be obtained as a solution to an ordinary differential equation of the following form,
$
\frac{d \bh_{\bx}(t)}{dt} = f(\bh_{\bx}(t), t, \pmb{\theta}_h)
$
Here, we assume the latent representations $\bh_{\bx}(t)$ is a function of time and changes continuously over time as defined by this ordinary differential equation.  
Solving the ODE requires one to provide an initial value $\bh_{\bx}(0)$ (initial value problem) and is typically considered as the input data $\bx$ or  a transformation  using a down-sampling block $d(\cdot)$. Given $\bh_{\bx}(0)$, hidden representation at some end-time $T$ can be obtained as
$\bh_{\bx}(T) = \bh_{\bx}(0) + \int_{0}^T f(\bh_{\bx}(t), t , \pmb{\theta}_h) $.
Since the direct computation of $\bh_{\bx}(T)$ is intractable,  numerical techniques such as Euler method or adaptive numerical techniques such as Dopri5 are used to obtain the final representation (ODESolve($f(\bh_{\bx}(t), t , \pmb{\theta}_h), \bh_{\bx}(0), 0, T$)). For e.g., Euler method is a single step method where $\bh_{\bx}(t)$ is updated sequentially until end-time $T$ with a step size $dt$. A particular step in the Euler method can be written as 
$
\bh_{\bx}(t+1) = \bh_{\bx}(t) + dt f(\bh_{\bx}(t), t , \pmb{\theta}_h)
$
We see that this is equivalent to the transformations performed in ResNet. On the other hand, adaptive numerical methods compute hidden representations at arbitrary times as determined by the error tolerance until the user specified end-time $T$. The end-time $T$ implicitly determines the number of transformations and consequently the depth of the network.  The hidden representation $\bh_{\bx}(T)$ is taken as  the final layer representation and is passed through a fully connected neural network to obtain the probability of predicting the output $y$, i.e. 
$p(y|\bx,T, \pmb{\theta}) = g_y(\bh_{\bx}(T), \pmb{\theta}_g).$
This predictive probability is then used with an appropriate loss function, for e.g., cross-entropy loss for classification, to obtain the final objective function. This is optimized to learn the parameters in the model using techniques such as adjoint sensitive method.   NODE based models provide a generalization performance close to ResNets with much smaller number of parameters and automates the model selection (depth selection) to some extent.

\section{Latent Time Neural Ordinary Differential Equations}
NODEs were found to be useful for many computer vision applications. However, their application to high-risk real-world problems such as healthcare and autonomous driving is limited by their lack of uncertainty modelling capability. We aim to develop efficient NODE models which can provide  good uncertainty estimates and make them amenable to such problems. We propose a novel approach, latent time neural ODE (LT-NODE), which is based on the idea of modelling  uncertainty through the uncertainty over end-time $T$. The proposed approach considers the hidden  representations at different end-times to obtain the predictive probability capable of modelling the model uncertainty or epistemic uncertainty. All the  representations from different end-times $T$ do not equally contribute to the predictive performance. Some of them will have a higher contribution than others. To account for this, we treat the end-time $T$ as a latent variable and learn a distribution over it from the data. To achieve this, we follow Bayesian learning principles~\cite{bishop06} where we define a prior distribution over $T$ and learn a posterior distribution $T$ from the data.  The prediction is done using the representations corresponding to the end-time sampled from the posterior over $T$.  The disagreements in the representations help to compute the model uncertainty. 
A side benefit of the proposed LT-NODE approach is that it automates the model selection over end time $T$. The posterior distribution over $T$ allows the model to learn the  end-time from the data. Moreover, our approach is generic and can be applied to model uncertainty with any recent NODE architecture.

The end-time $T$ associated with NODE takes a positive real value and we would like to evaluate the representations at arbitrary times in the positive real valued interval to compute uncertainty. This makes NODEs challenging and  different from discrete depth neural networks and we need an appropriate distribution which allows this.   This motivates us to  use Gamma distribution whose support is $(0,\infty)$ as the prior over $T$  with shape and rate parameters being $\alpha_p$ and $\beta_p$ respectively, thus $p(T |\alpha_p,\beta_p) = \text{Gamma}(T|\alpha_p, \beta_p)  = \frac{\beta_p^{\alpha_p}}{\Gamma(\alpha_p)} T^{\alpha_p - 1} e^{-\beta_p T}$, where  $\Gamma(\alpha_p)$ is gamma function.   Gamma distribution can be useful to model the end-times  as it is more flexible than exponential distribution and can  place its probability density over end-times in any arbitrary region. The likelihood of modelling outputs $\by$ given a value for the end-time $T$ and inputs $X$ is denoted  as  $p(\by|T,X,\pmb{\theta}) = \prod_{i=1}^N p(y_i|T,\bx_i,\pmb{\theta})$  where  $p(y_i|T,\bx_i,\pmb{\theta}) = g_{y_i}(\bh_{\bx_i}(T), \pmb{\theta}_g)$. Given the likelihood and the prior, the posterior over the latent variable $T$ can be computed using the Bayes theorem~\cite{bishop06} as 
\begin{equation}
    p(T|\by,X;\pmb{\theta}) = \frac{p(\by|T,X;\pmb{\theta})p(T|\alpha_p,\beta_p)}{\int_{0}^{\infty}p(\by|T,X;\pmb{\theta})p(T|\alpha_p,\beta_p)dT}
    \label{eq:posterior}
\end{equation}
However,  the posterior cannot be computed in a closed form as  the end-time $T$  appears as  a complex non-linear function in the likelihood. Consequently, the marginal likelihood term in the denominator of \eqref{eq:posterior} also cannot be computed.  Hence, we resort to approximate inference techniques such as the variational inference~\citep{blei17} to obtain an approximate posterior over the $T$. 

\subsection{Variational Inference}
In our approach,  we choose Gamma distribution as the variational posterior over $T$ (due to  positive real valued $T$) with  variational parameters $\alpha_q$ and $\beta_q$, thus $q(T|\alpha_q,\beta_q) = \text{Gamma}(T|\alpha_q, \beta_q)$. We derive the variational lower bound or evidence lower bound (ELBO) for our setting as follows.
\begin{multline}
   \log(p(\by|X;\pmb{\theta}) \geq \mathbb{E}_{q(T|\alpha_q,\beta_q)}[\log(p(\by|T,X;\pmb{\theta}))] - \\ \mathbb{KL}((q(T|\alpha_q,\beta_q)||p(T|\alpha_p,\beta_p))).
    \label{eq:lower_bound}
\end{multline}
We learn the  variational posterior parameters by maximising the lower bound.  The $\mathbb{KL}$ term in \eqref{eq:lower_bound} can be computed in closed form~\citep{bauckhage2014computing} as
\begin{multline}
    \alpha_q\log \beta_q - \alpha_p\log \beta_p + log(\Gamma(\alpha_p)) - log(\Gamma(\alpha_q)) \\+  (\psi(\alpha_q)-log\beta_q )(\alpha_q - \alpha_p) + \frac{\Gamma(\alpha_q + 1)}{\Gamma(\alpha_q)}\frac{\beta_p}{\beta_q} - \alpha_q  \nonumber
\end{multline}
where $\psi$ is a digamma function.
We approximate the computation of the expectation term in the ELBO by discretizing the space of $T$ into a uniform grid and use $S$ samples of $T$ from the uniform grid to approximate the expectation as
\begin{equation}
\begin{aligned}
    & \mathbb{E}_{{q(T|\alpha_q,\beta_q)}}[log(p(\by|T,X,\pmb{\theta}))] \\ 
    & = \sum_{i=1}^{N} \sum_{s=1}^{S} \log(p(y_i|T_s,\bx_i,\pmb{\theta}))  q(T_s|\alpha_q,\beta_q),
   \label{eq:MC}
\end{aligned}
\end{equation}
where, $T_s \sim Uniform(T|a,b)$. We decided to use the uniform grid approximation rather than Monte Carlo approximation because of two reasons. Firstly, the latent variable $T$ is a scalar quantity  and consequently this approach will not suffer from the sampling inefficiency typically associated with high dimensional variables which motivate the use of Monte Carlo sampling techniques. Secondly, uniform grid approximation allows us to consider the variational distribution $q(T|\alpha_q,\beta_q)$ explicitly in the objective function and makes sampling independent of the parameters to be estimated. This will ease the estimation of variational parameters using gradient descent. We maximize ELBO and back-propagate the gradients to estimate both the variational  and  model parameters ($\pmb{\theta}$). 

Let $\bar{S} = \{T_1,T_2, \ldots, T_S\}$ be the $S$ end-times sampled  from the uniform distribution. During forward propagation, intermediate feature vectors are computed using an adaptive numerical technique such as  Dopri5~\cite{kimura2009dormand})  until $T_S$. 
The feature vectors at $T_i \in \bar{S}$ 
 obtained using the adaptve numerical technique and interpolation  are used  to compute the approximate log  probability  of training samples \eqref{eq:MC} and consequently in \eqref{eq:lower_bound} to obtain ELBO. We note that this can be computed efficiently by ordering the sampled times (in ascending order) and obtaining the features vectors at these times in a single forward pass.   

\begin{algorithm}[H]
\begin{algorithmic}
\STATE \%Sample $S$ end-times from the variational posterior $q(T|\alpha_q,\beta_q)$, Initialize $\bar{S} = \{\}$ \\
\textbf{while} $|\bar{S}| \leq S $ \textbf{do}\\
\hspace{0.3cm} Sample $T_s \sim q(T|\alpha_q,\beta_q)$ \\
\hspace{0.3cm} $\bar{S} = \bar{S} \cup T_s $ \\
Sort $\bar{S}$ in increasing order \\
Transform input using the downsampling: $\bh_\bx(0) = d(\bx)$\\
\textbf{initialize} : $t=0$ , $\mathbf{prob\_vec} = \mathbf{0}$ \\
\hspace{0cm} \textbf{for} $T_s$ in $\bar{S}$\\
\hspace{0.3cm}  $\bh_\bx(T_s) =\text{ODESolve}(f,\bh_\bx(t), t,T_s)$\\ 
\hspace{0.3cm}  $t = T_s$\\
\hspace{0.3cm} \textbf{for} $y = 1,\ldots,C $\\
\hspace{0.6cm}  $\mathbf{sample\_prob\_vec}(y) = g_{y}(\bh_{\bx}(t), \pmb{\theta}_g)$\\
\hspace{0.3cm}  $\mathbf{prob\_vec} = \mathbf{prob\_vec}+$   $\mathbf{sample\_prob\_vec}$ \\
\hspace{0.0cm} \textbf{return} $\mathbf{prob\_vec} = \frac{\mathbf{prob\_vec}}{S}$
 \caption{Forward pass in LT-NODE, computing predictive probability for datapoint $\bx$. }
 \label{alg:adjoint}
 \end{algorithmic}
\end{algorithm}

\begin{figure}
\begin{center}
\includegraphics[scale=0.30]{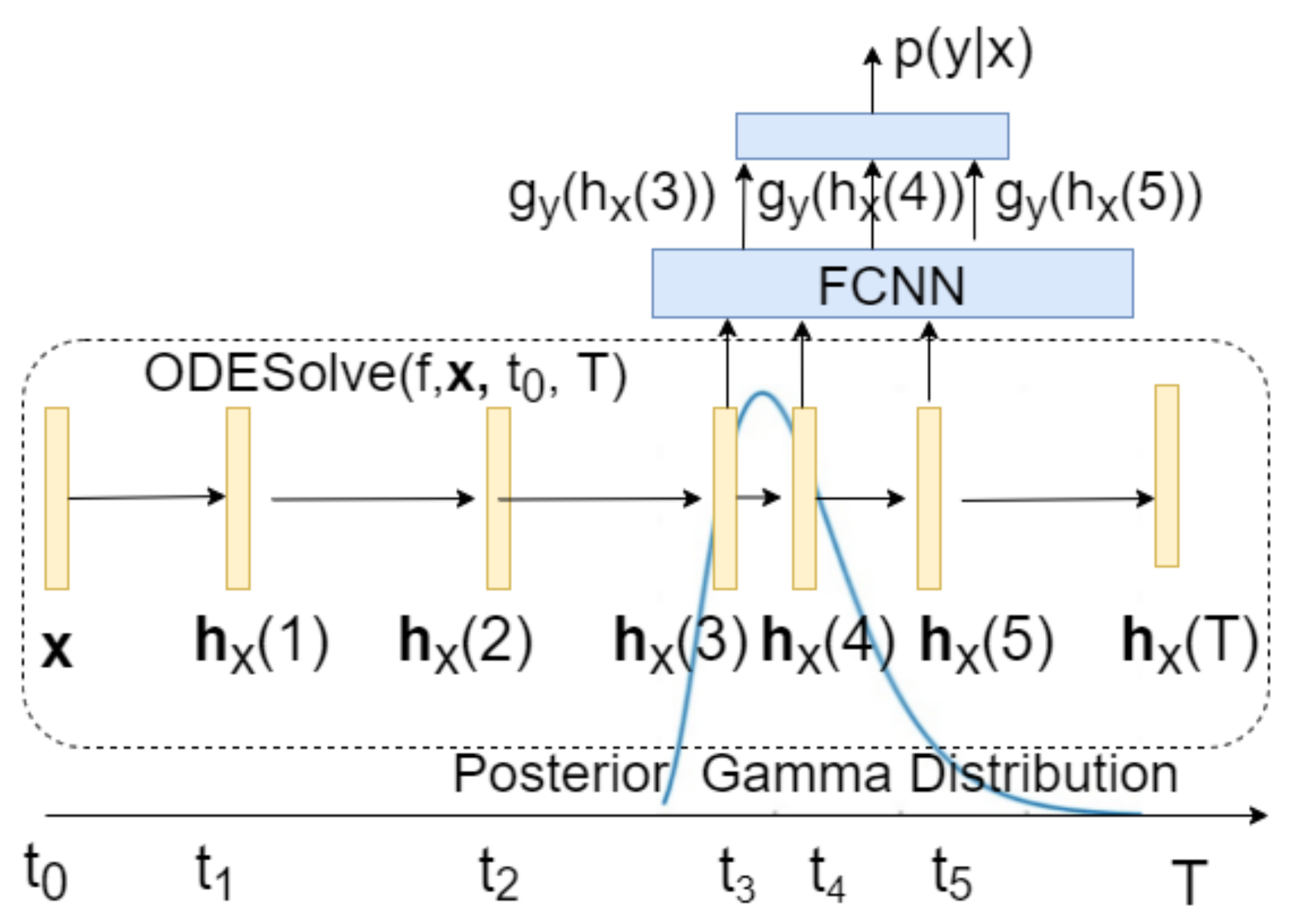}
\end{center}
\caption{ Representation of the LT-NODE. Posterior over end-times is Gamma distributed, and assume $t_3$, $t_4$, and $t_5$ are the end-times sampled from the Gamma. The representations at these times are passed through the FCNN and the output is averaged  to get the final predictive probability. } 
\label{fig:ltnode}
\vspace{-1em}
\end{figure}

The model parameters and variational parameters learnt by maximizing ELBO are used to predict the test data. First, we sample the  end-times from the learnt variational posterior $q(T|\alpha_q, \beta_q)$. The sampled end-times are ordered, and predictions are  done efficiently using  a single forward pass through the model in a similar manner as discussed for training. We compute the predictive probability  of a test data point $\bx$ to be classified to a class $y$ as      
   $\frac{1}{S} \sum_{s=1}^{S} p(y|T_s,\bx,\pmb{\theta})$, where $T_s \sim q(T|\alpha_q,\beta_q)$. 
LT-NODE  provides good uncertainty estimates with only 2 additional parameters ($\alpha_q$ and $\beta_q$). A schematic representation and a detailed algorithm of the proposed LT-NODE are shown in  Figure~\ref{fig:ltnode} and Algorithm~\ref{alg:adjoint}.

\section{Adaptive Latent Time Neural Ordinary Differential Equations}

LT-NODE computes a posterior distribution over end-time $T$ which helps to  model uncertainty as well as aid in   model selection. However, end-time $T$ is treated as a global latent variable and the distribution over $T$ is assumed to be same for all the data points.
Though this gives a good uncertainty estimate, NODE modelling capability can be  improved by considering the end-time to be different across different data points. 
~\citet{massaroli2020dissecting} showed that a NODE with input specific depth will be able to model complex problems such as solving \textit{ concentric annuli} and \textit{reflection} tasks.  To improve the modelling capability, we propose a variant, adaptive latent time NODE (ALT-NODE), which allows each data point to have input specific distribution over end-time. 

In ALT-NODE, we assume that every data point is associated with a latent variable $T_i$ denoting the end-time associated with the data point. We assume the same Gamma prior over $T_i$ with parameters $\alpha_p$ and $\beta_p$ as before. The likelihood $p(y_i|T,\bx_i,\pmb{\theta})$  is also defined as in LT-NODE. Here, we will be learning a separate posterior distribution over $T_i$ associated with each data point. As in LT-NODE, the posterior cannot be computed tractably, and we resort to variational inference to obtain an approximate posterior. For ALT-NODE, we associate a separate variational posterior $q(T_i)$ with each $T_i$.
Due to the nature of $T_i$, we assume $q(T_i)$ to be Gamma distributed with parameters $\alpha_{qi}$ and $\beta_{qi}$. Treating the variational parameters as free form distribution can lead to a few drawbacks. Firstly, the number of  variational parameters to learn increases linearly with number of data points and can become costly when number of data points is high.  Secondly, it does not allow us to perform inference over  new data points. 

We use an amortized variational inference~\citep{Gershman2014AmortizedII,kingma2013auto} approach to address these drawbacks. The amortized VI assumes the variational parameters associated with the $T_i$ can be obtained as a function of the input data points. It introduces an inference network, typically a parametric function such as neural networks, which can predict the variational parameters from the input data. Now, instead of learning the variational parameters, one can learn the parameters of the inference network from the variational lower bound.  Learning of the inference network allows  statistical strength to be shared across data points and helps in predicting the variational parameters for a new data point. Therefore, we introduce an inference network  $r(\bx_i; \phi)$  which predicts the variational parameters $\alpha_{qi}$ and $\beta_{qi}$ associated with $T_i$. Consequently,  we denote the  variational distribution over $T_i$, to be parameterized by the inference network parameters $\phi$ and is conditioned on $\bx_i$, i.e., $q(T_i| \bx_i, \phi)$. We learn the inference network parameters $\phi$ and model parameters $\theta$ by maximizing  the variational lower bound for ALT-NODE which is  derived as follows
\vspace{-0.25cm}
\begin{multline}
     \sum_{i=1}^{N}  [ \mathbb{E}_{q(T_i|\bx_i, \phi)}[\log(p(y_i|T_i,\bx_i;\pmb{\theta}))] - \\ \mathbb{KL}((q(T_i|\bx_i, \phi)||p(T_i|\alpha_p,\beta_p))) ].
    \label{eq:lower_bound_atunode}
\end{multline}
We follow the approximation used in LT-NODE to evaluate the expectation term in the objective function \eqref{eq:lower_bound_atunode}. The KL divergence term can be obtained in closed form as before, but the variational parameters 
is a  function of the inference network $r(\bx)$ parameterized by $\phi$. We developed an efficient approach to perform single forward pass computation through the ALT-NODE similar to LT-NODE for training and prediction~\footnote{Details of the approach in supplementary}.  

\begin{figure*}[t]
\begin{center}     
\subfigure[LT-NODE]{\includegraphics[scale=0.2]{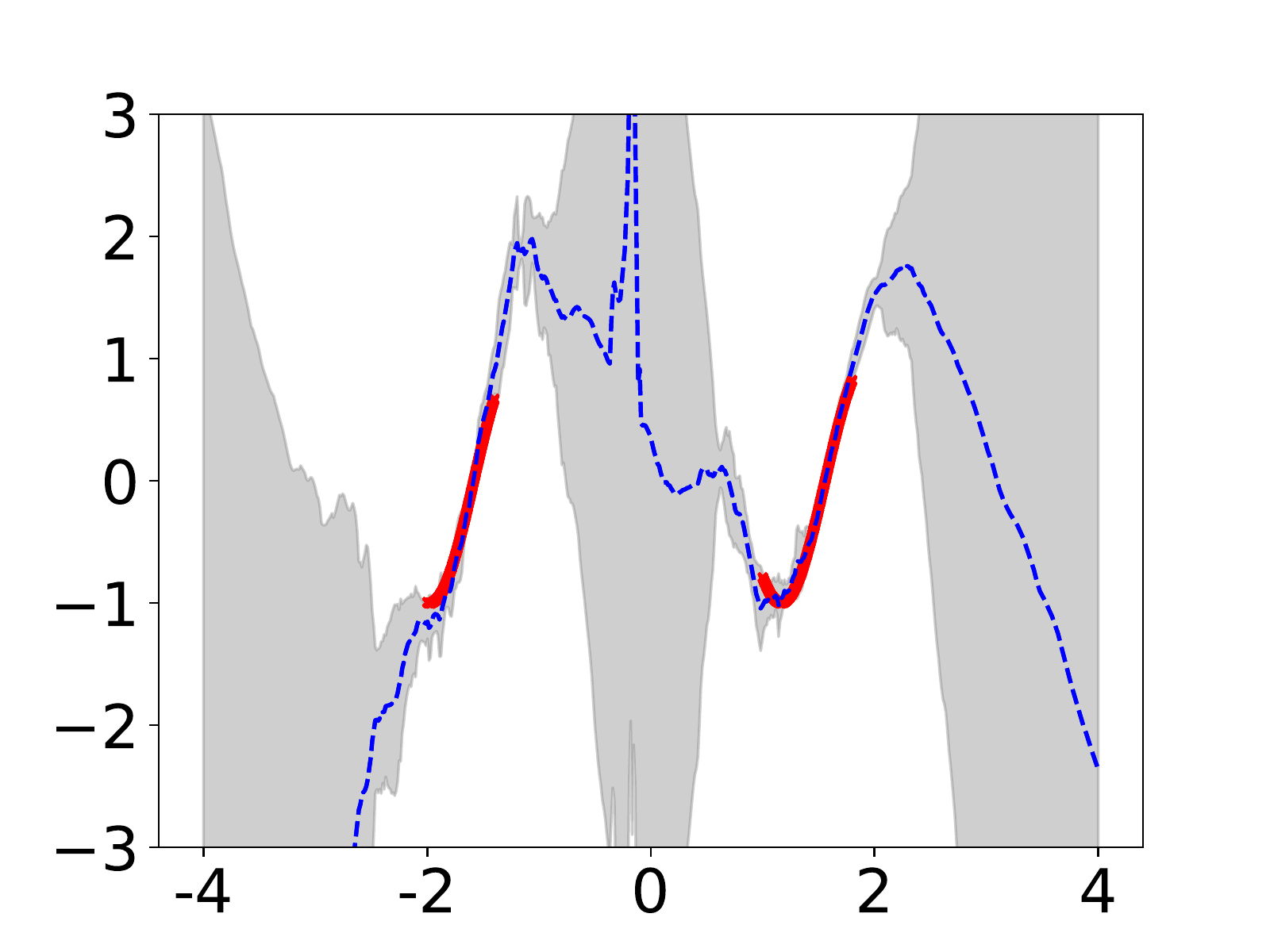}}
\subfigure[ALT-NODE]{\includegraphics[scale=0.2]{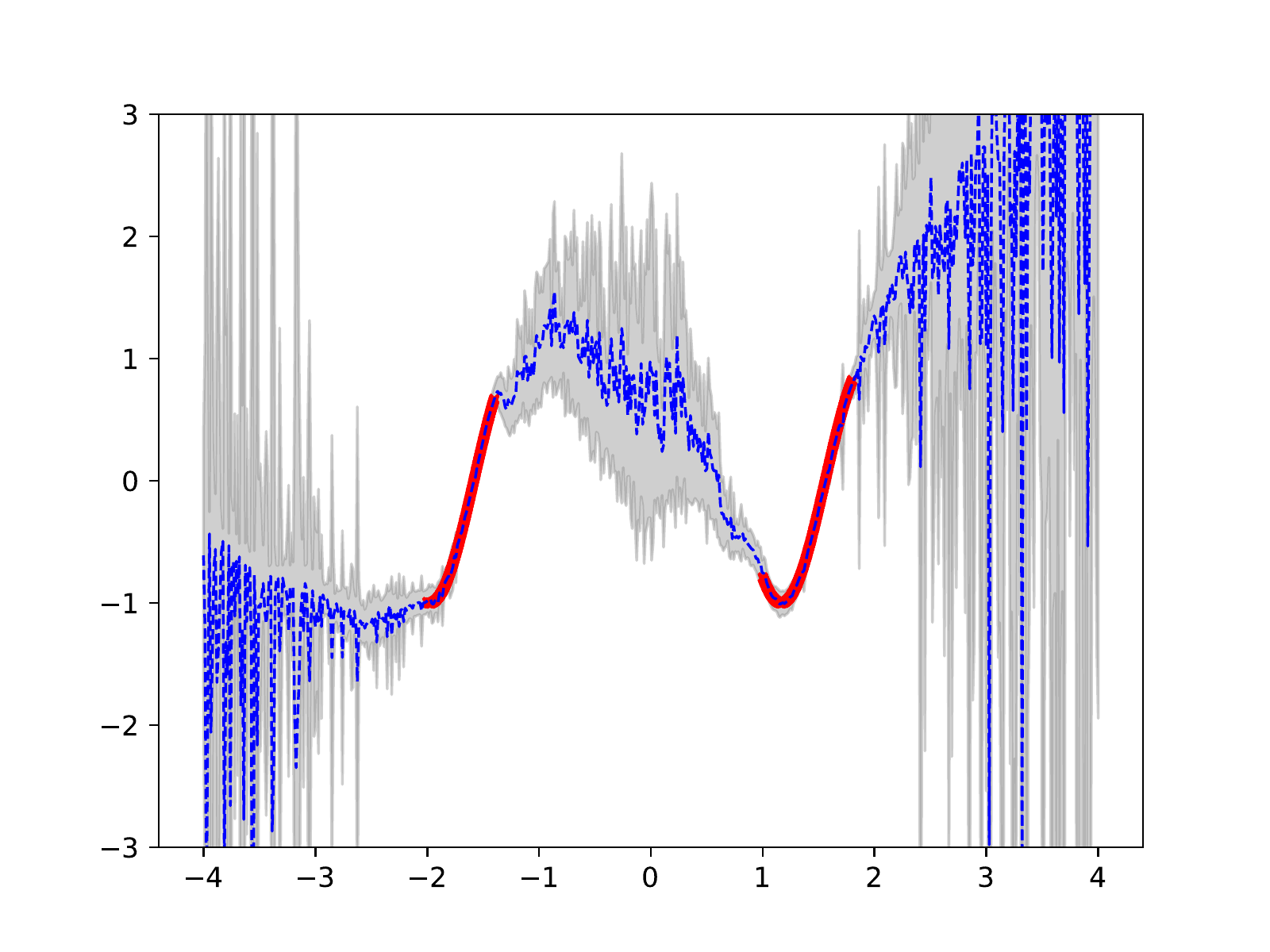}}
\subfigure[SDE-Net]{\includegraphics[scale=0.2]{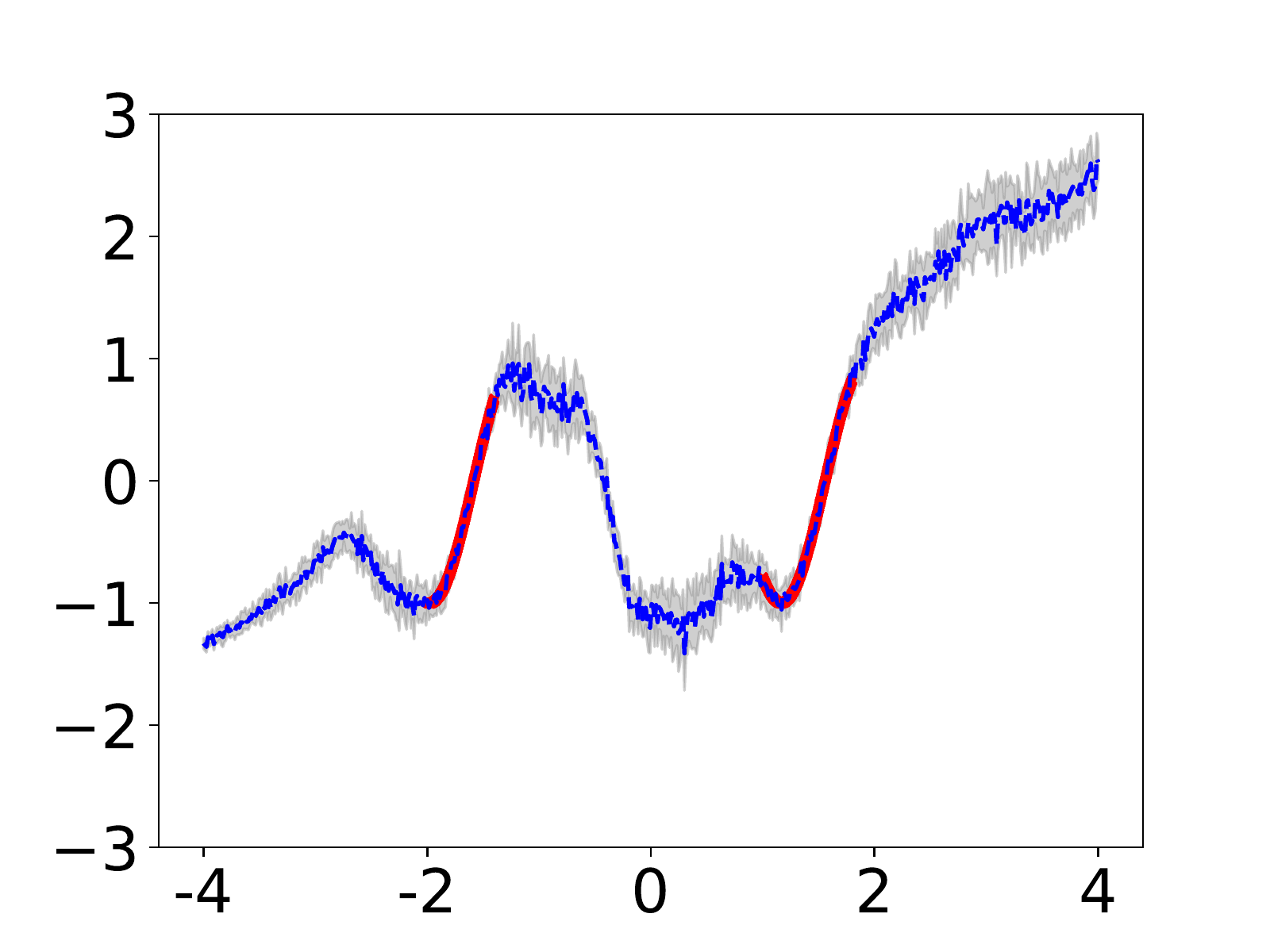}}
\subfigure[NODE-GP]{\includegraphics[scale=0.2]{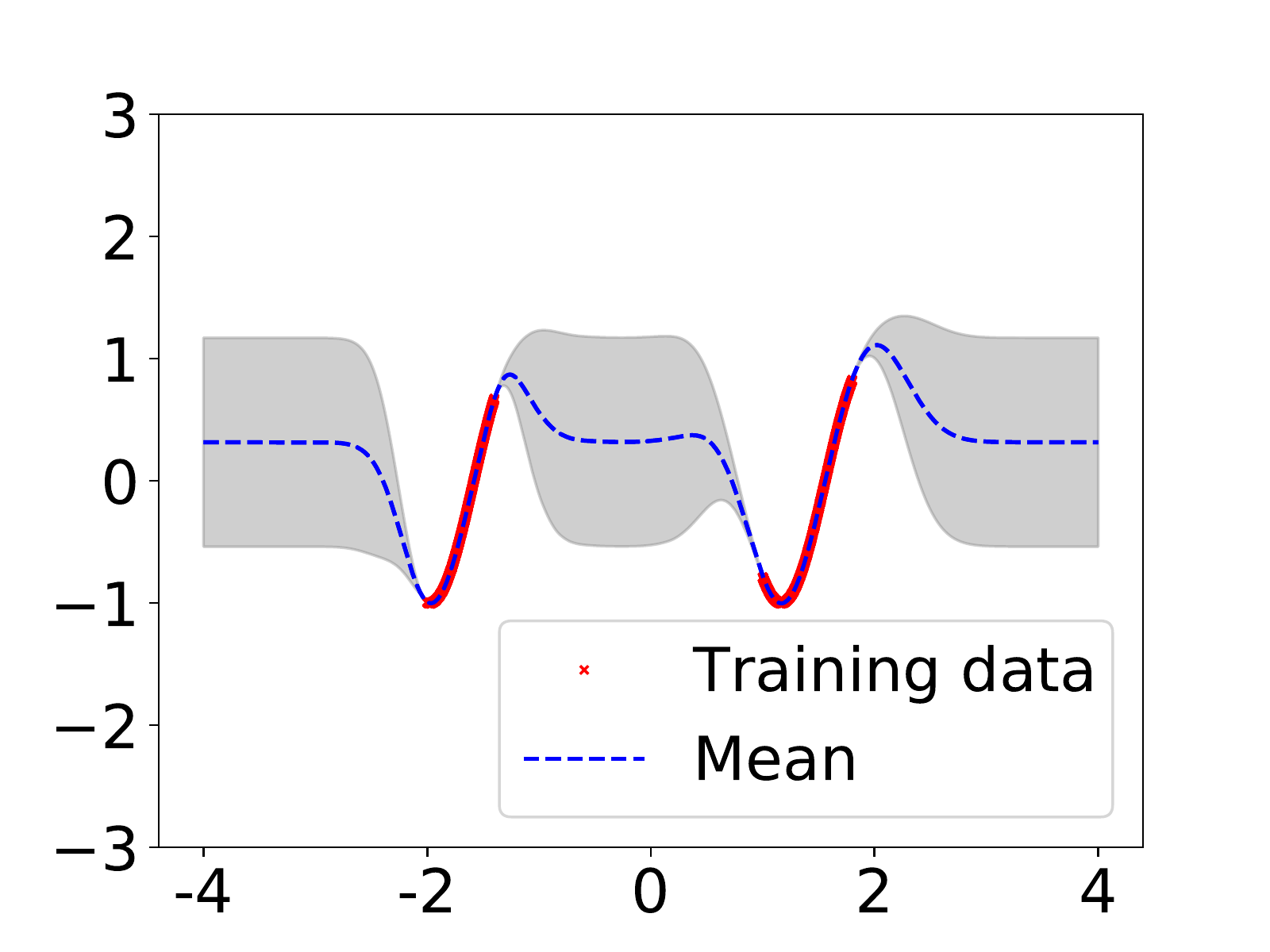}}
\subfigure[Uni-NODE]{\includegraphics[scale=0.20]{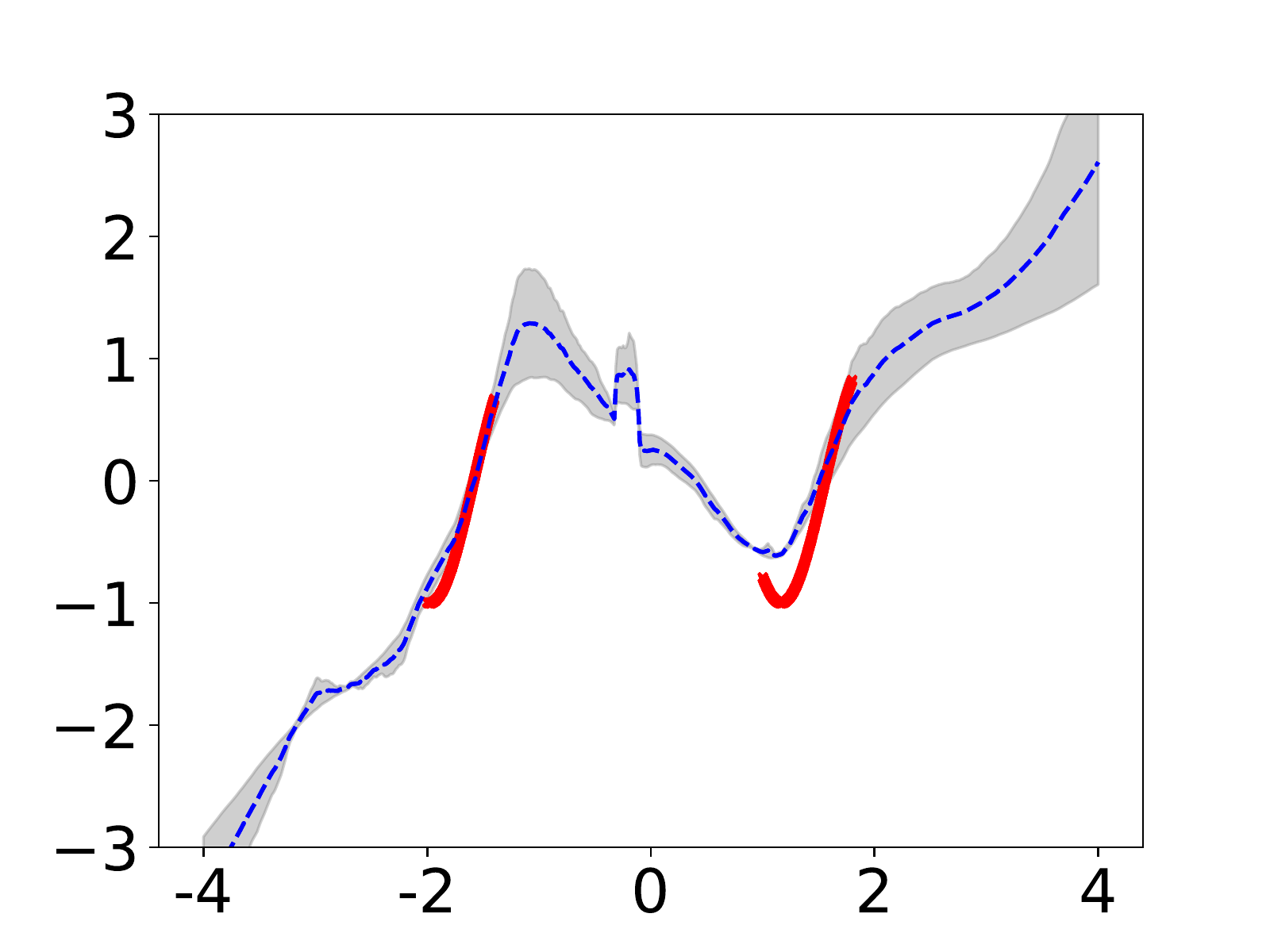}}
\end{center}
\vspace{-1em}
\caption{Results on  1-D synthetic regression  data~\citep{foongbetween}. Mean prediction is denoted by dotted blue line and shaded region represents mean $\pm$ std. deviation. We also provide average entropy (E) computed in the OOD interval $(-0.5,0.5)$. (a) LT-NODE (E:$2.42$) exhibits a good uncertainty modelling capability followed by (d) NODE-GP (E:$1.22$) and (b) ALT-NODE (E:$1.17$). (c) SDE-Net (E:$-0.31$) exhibits some uncertainty, but it remains stagnant in the OOD regime. (e) Uni-NODE (E:$-0.65$) exhibits some uncertainty but does not fit the data well.} 
\label{fig:synthetic}
\vspace{-1em}
\end{figure*}
\section{Experiments}
We conduct experiments to evaluate the  uncertainty  and robustness modelling  capabilities of the proposed approaches\footnote{https://github.com/srinivas-quan/LTNODE}, LT-NODE and ALT-NODE using synthetic and real-world data sets. The approaches are compared against standard NODE~\citep{node} and  baselines which were recently proposed to model uncertainty in the NODE models, such as NODE-GP~\citep{anumasa2021improving}  and SDE-Net~\citep{kong2020sdenet}. We also consider a baseline Uni-NODE which does not learn any posterior over $T$ but only considers randomness over $T$ by sampling it from a uniform distribution during training and testing. This baseline is a variant of~\cite{ghosh20} where the model considered a noisy end-time during training but not during testing. 
\subsection{Synthetic Data Experiments}
We consider a 1D synthetic regression dataset~\citep{foongbetween} to demonstrate the uncertainty modeling capability of the proposed models~\footnote{We use Gamma($2,0.5$) as prior and  learn the approximate variational posterior by maximizing ELBO, for e.g., LT-NODE learnt Gamma($1.27,0.98$) as the approximate posterior on the synthetic data, more details in the  supplementary.}. This dataset contains two disjoint clusters of training points. We expect the models to exhibit high variance in-between and away from these training data points. Figure~\ref{fig:synthetic} provide the predictive mean and standard deviation obtained with the proposed models and baselines. In this 1-D regression problem,   LT-NODE model as shown in Figure~\ref{fig:synthetic}(a) captures the uncertainty well with 
high variance on in-between and  away data points on the left, and the variance grows smoothly. Infact, it is found to have highest in-between variance. ALT-NODE in Figure~\ref{fig:synthetic}(b) also captures the uncertainty well, and  due to the input conditioned posterior it is able to fit and learn the trends in data  better than LT-NODE.  We find that SDE-Net in Figure~\ref{fig:synthetic}(c) exhibits some uncertainty but the variance does not increase as we move away from training data regime.
NODE-GP  in Figure~\ref{fig:synthetic}(d) gives a good uncertainty  modelling capability with a high variance on both in-between and away OOD data. But we show later that on high-dimensional image data, NODE-GP fails to model the uncertainty due to the inability of GPs to model high dimensional data. 
We also conducted experiments to demonstrate the importance of learning a posterior distribution over $T$. In Figure~\ref{fig:synthetic}(e), 
we can observe that Uni-NODE  which does a random sampling of $T$ was not able to fit the data unlike other models but having a randomness over $T$ provided some uncertainty modeling capability in the away data region.

\begin{figure*}[t]
\begin{center}     
\subfigure[LL (CIFAR10)]{\includegraphics[scale=0.2]{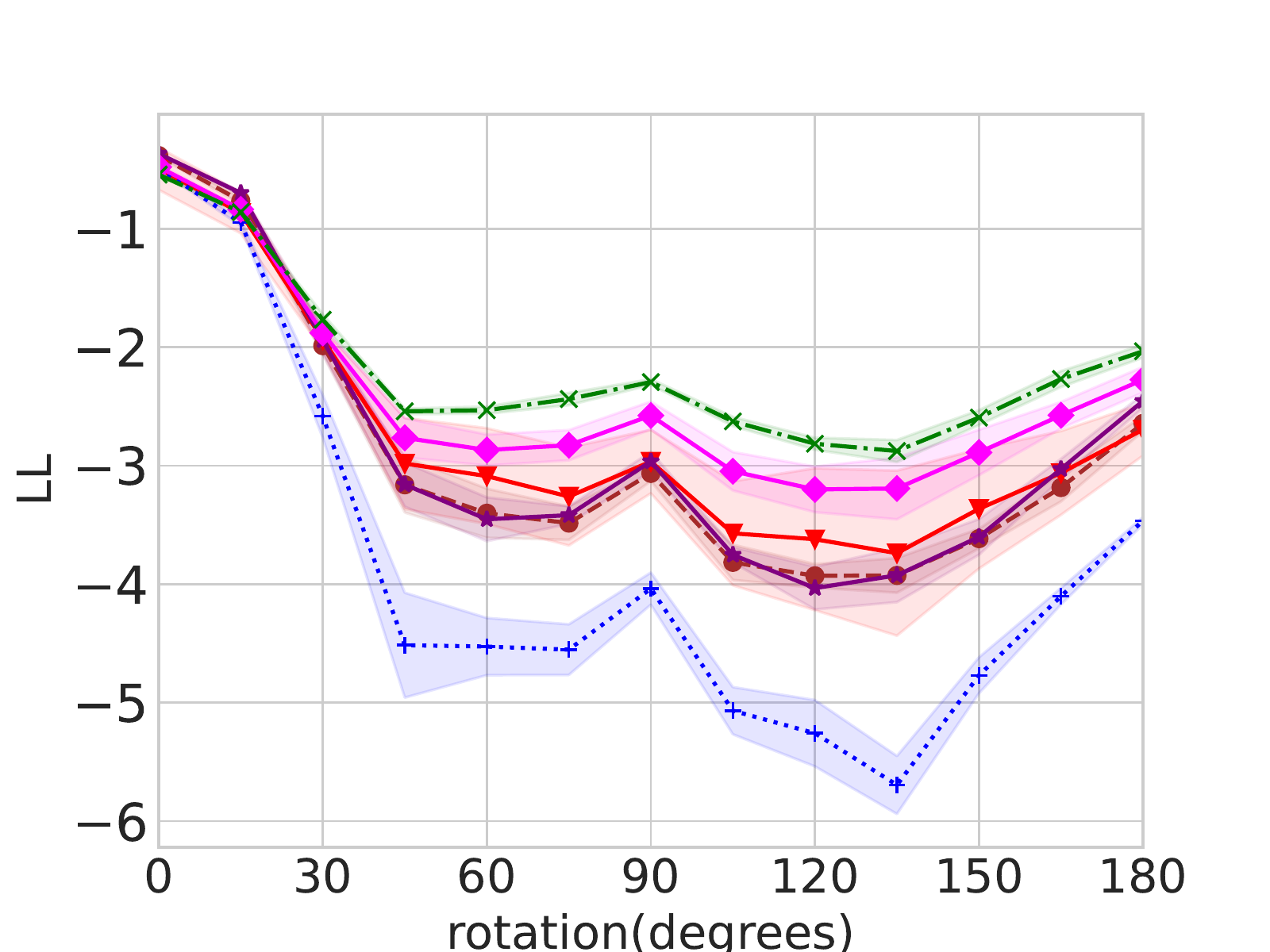}}
\subfigure[Brier (CIFAR10)]{\includegraphics[scale=0.2]{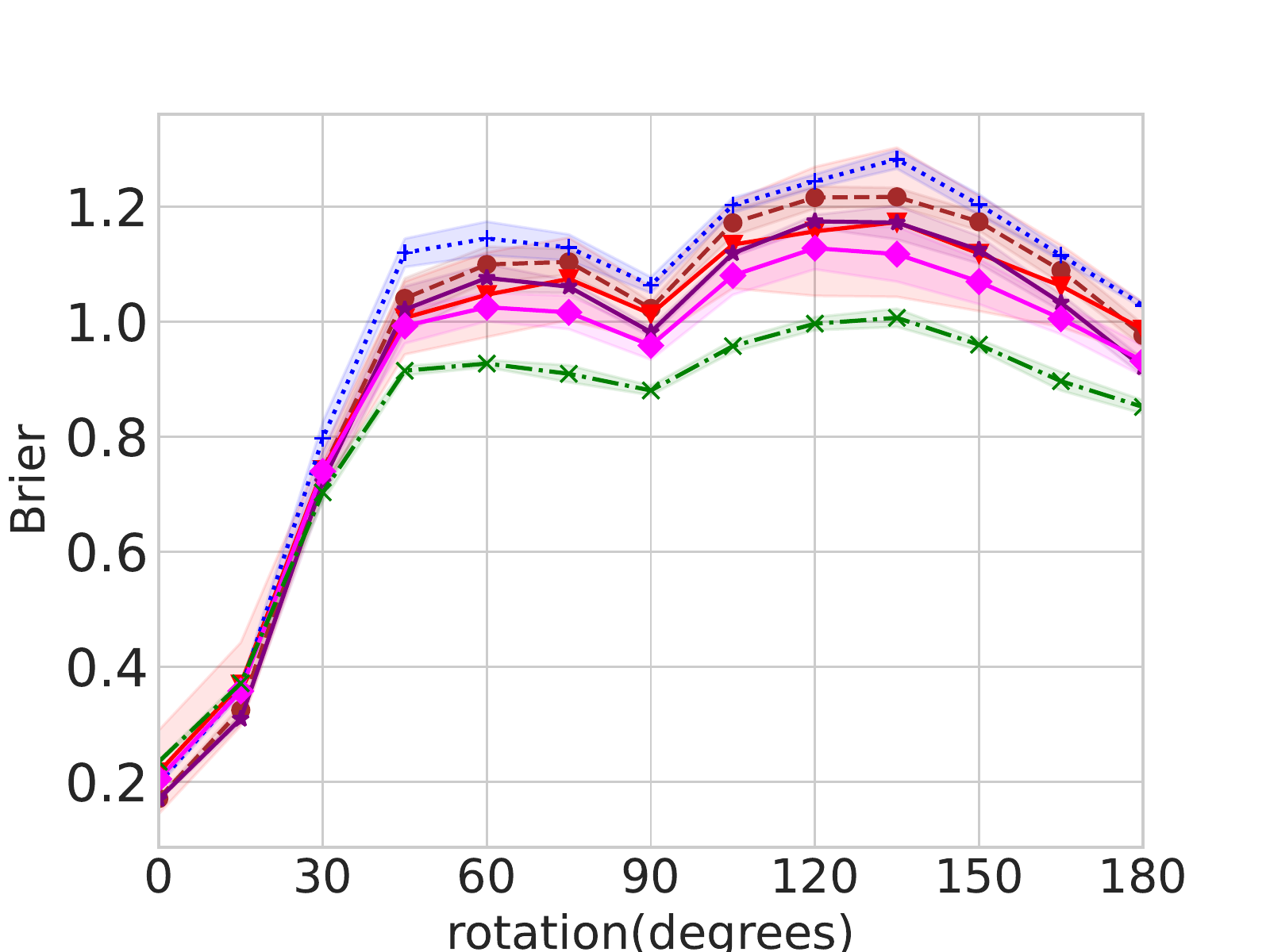}}
\subfigure[ECE (CIFAR10)]{\includegraphics[scale=0.2]{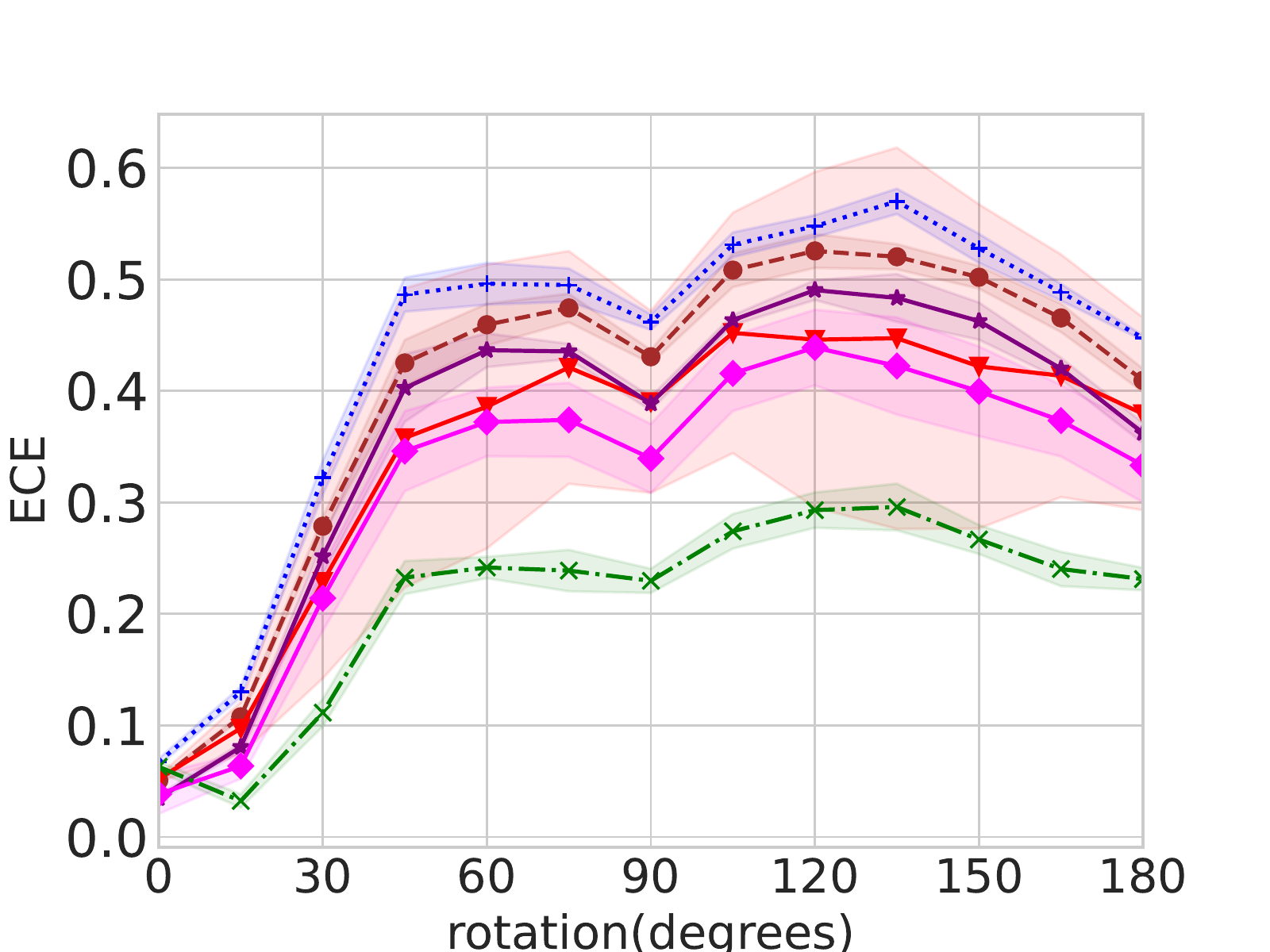}}
\subfigure[Accuracy vs Confidence (Rotated CIFAR10)]{\includegraphics[scale=0.2]{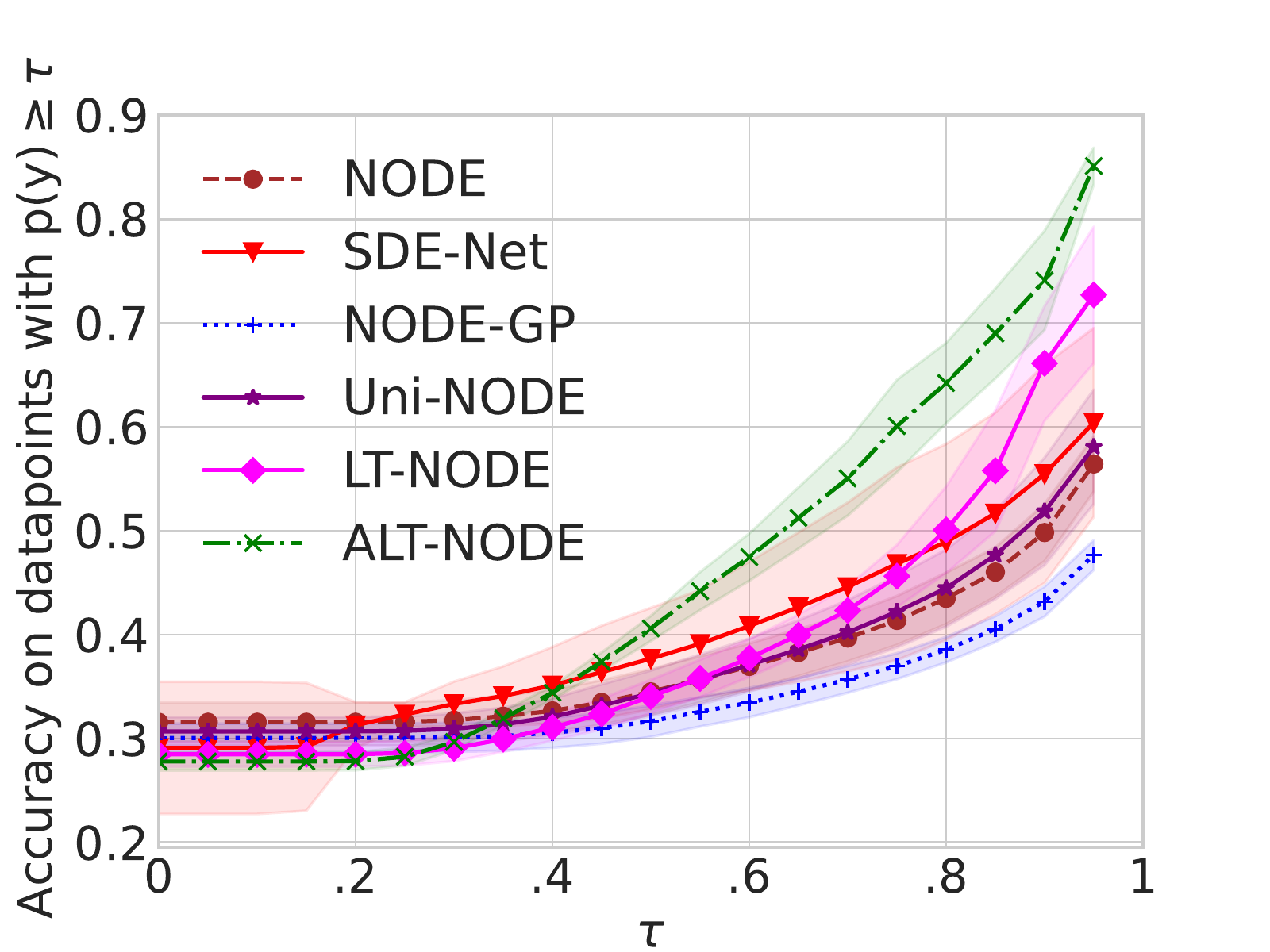}}
\subfigure[Count vs Confidence (Rotated CIFAR10)]{\includegraphics[scale=0.2]{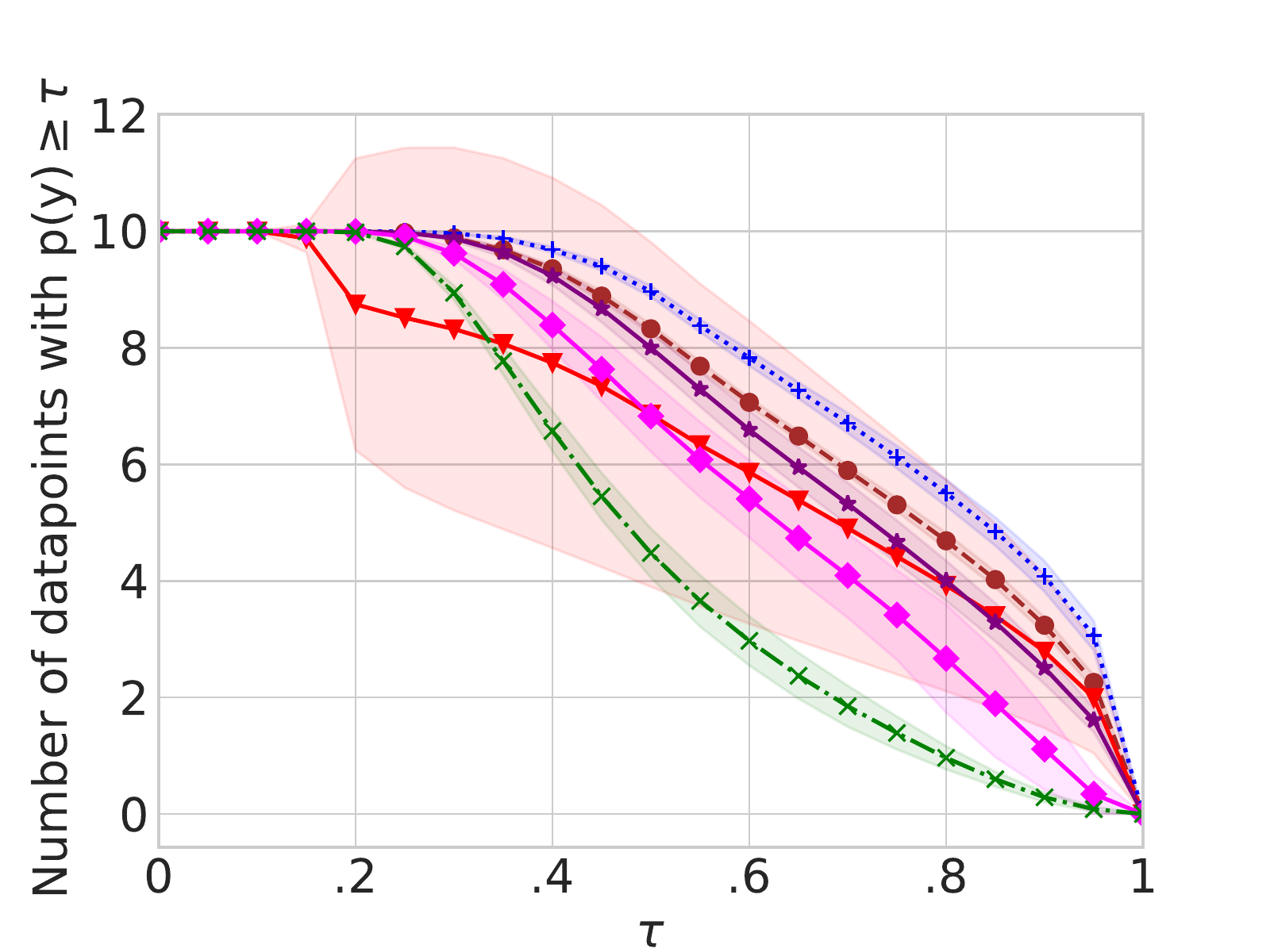}}
\subfigure[LL (SVHN)]{\includegraphics[scale=0.2]{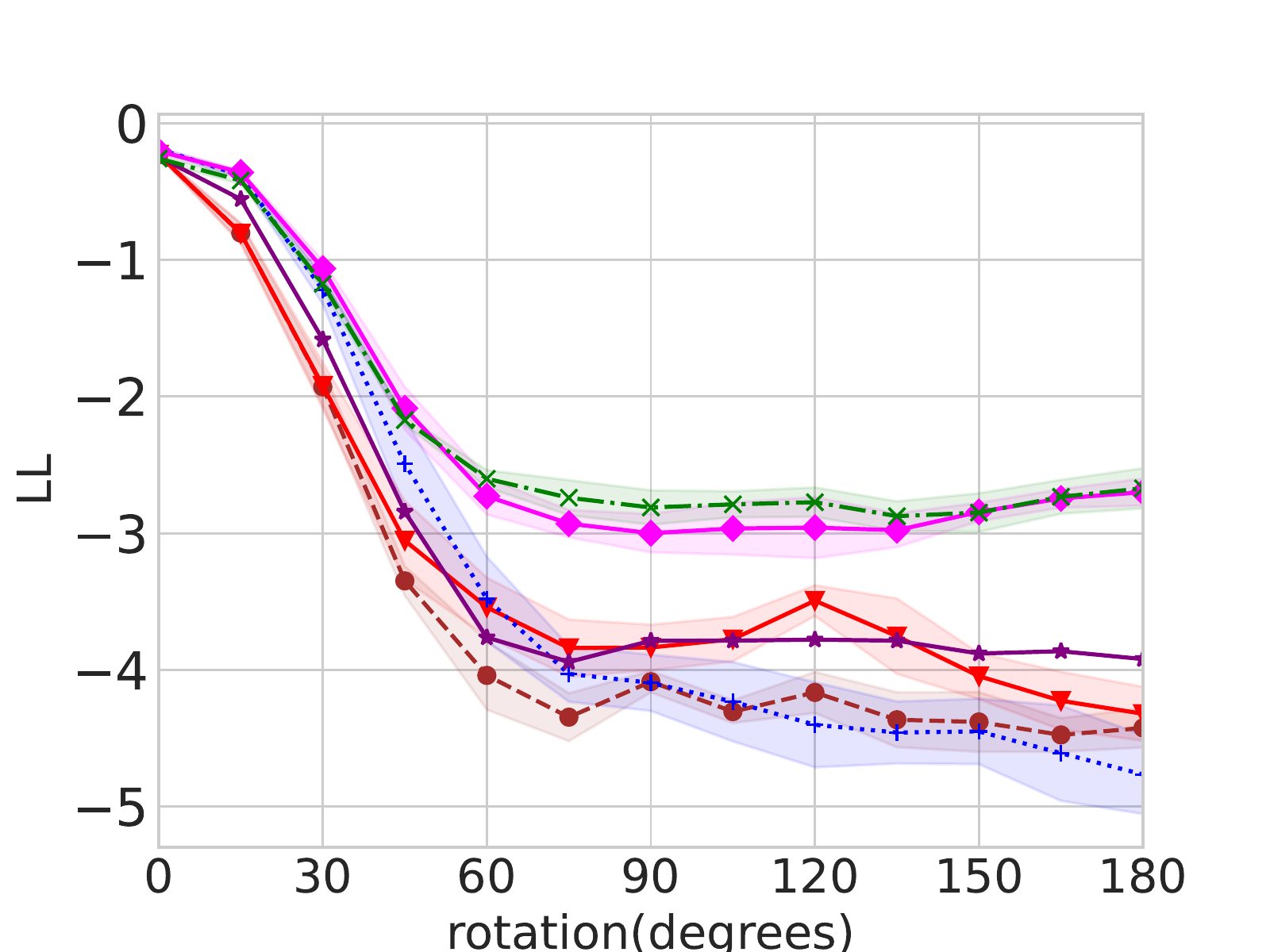}}
\subfigure[Brier (SVHN)]{\includegraphics[scale=0.2]{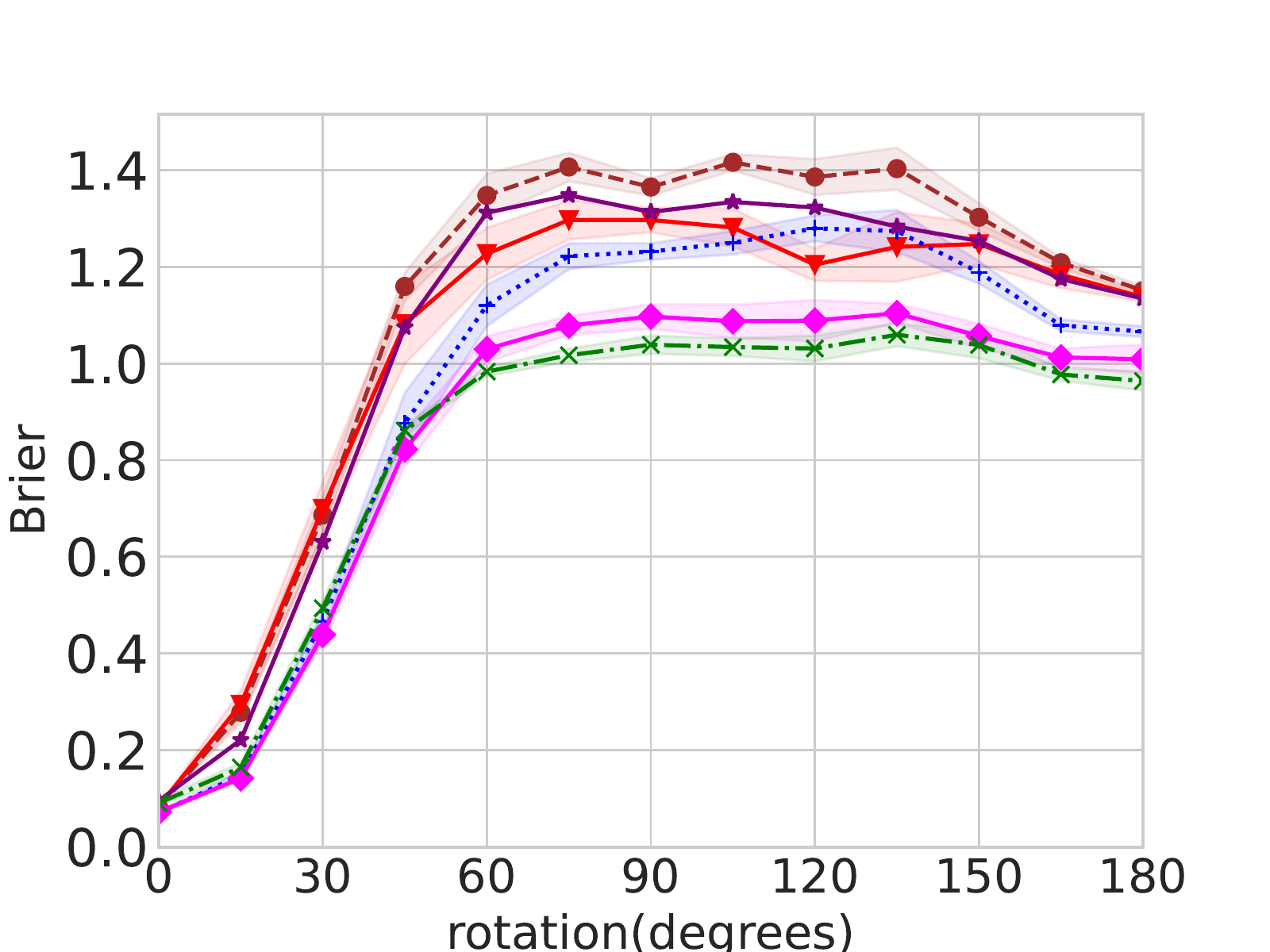}}
\subfigure[ECE (SVHN)]{\includegraphics[scale=0.2]{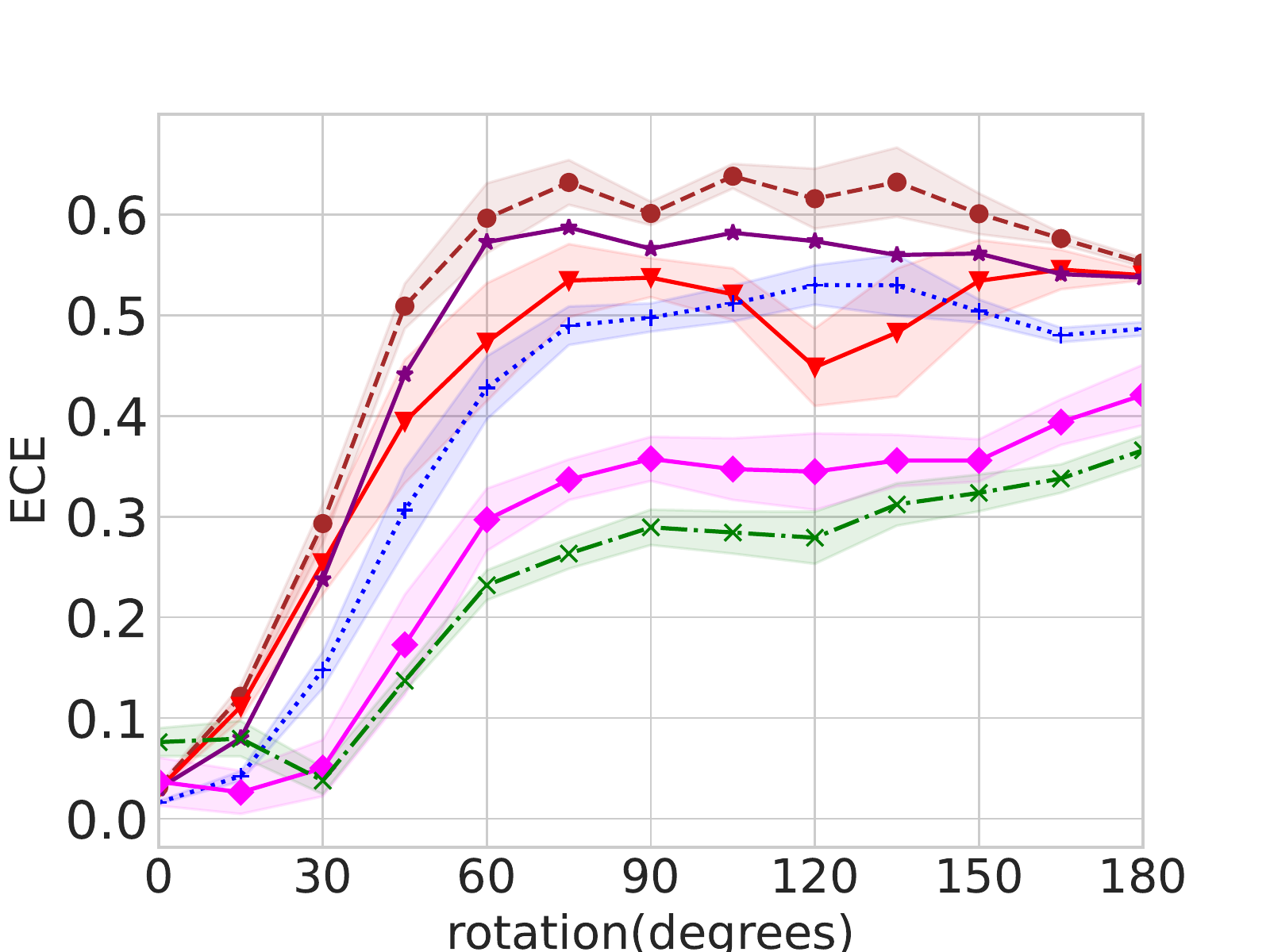}}
\subfigure[Accuracy vs confidence (Rotated SVHN)]{\includegraphics[scale=0.2]{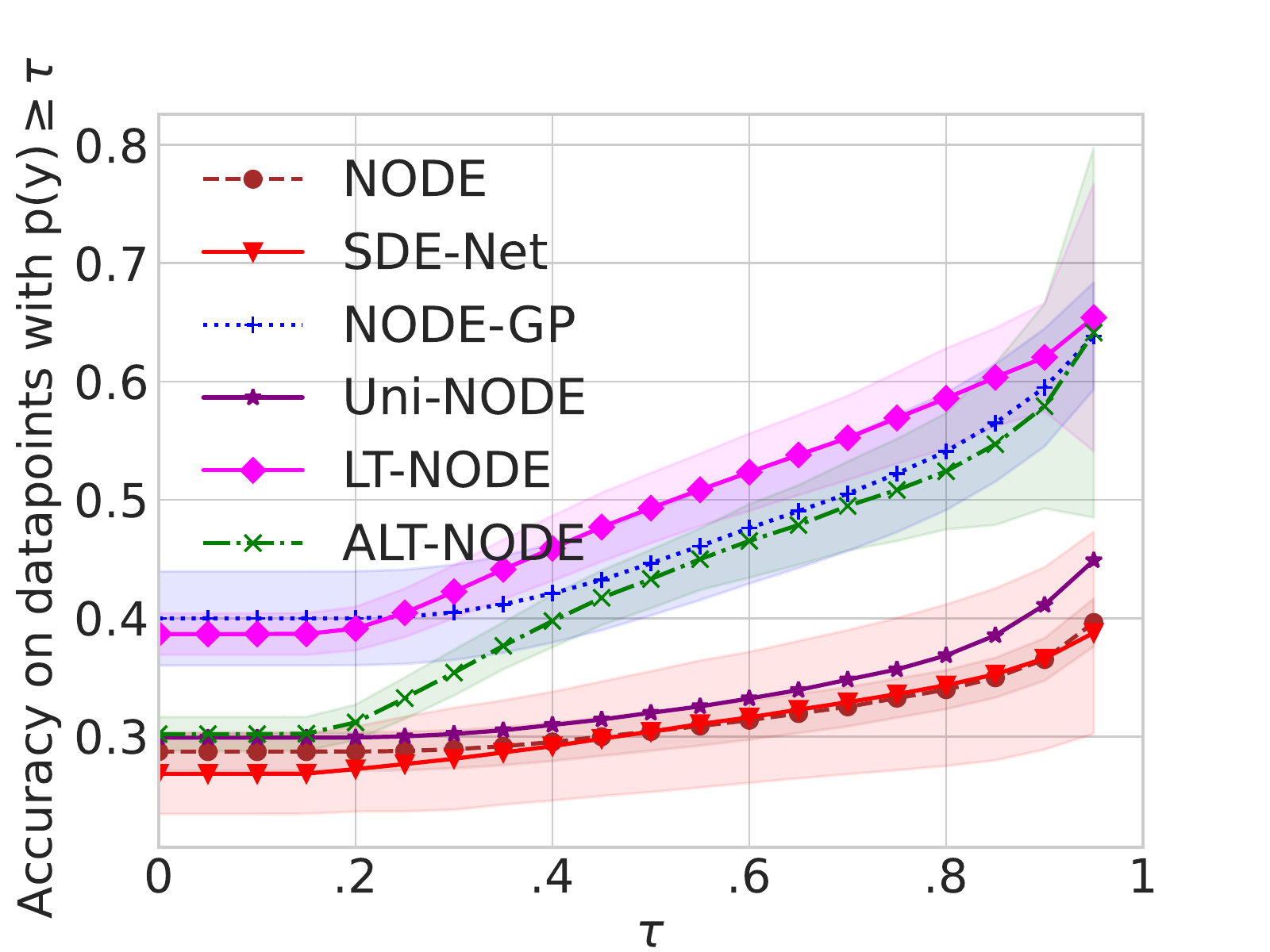}}
\subfigure[Count vs Confidence (Rotated SVHN)]{\includegraphics[scale=0.2]{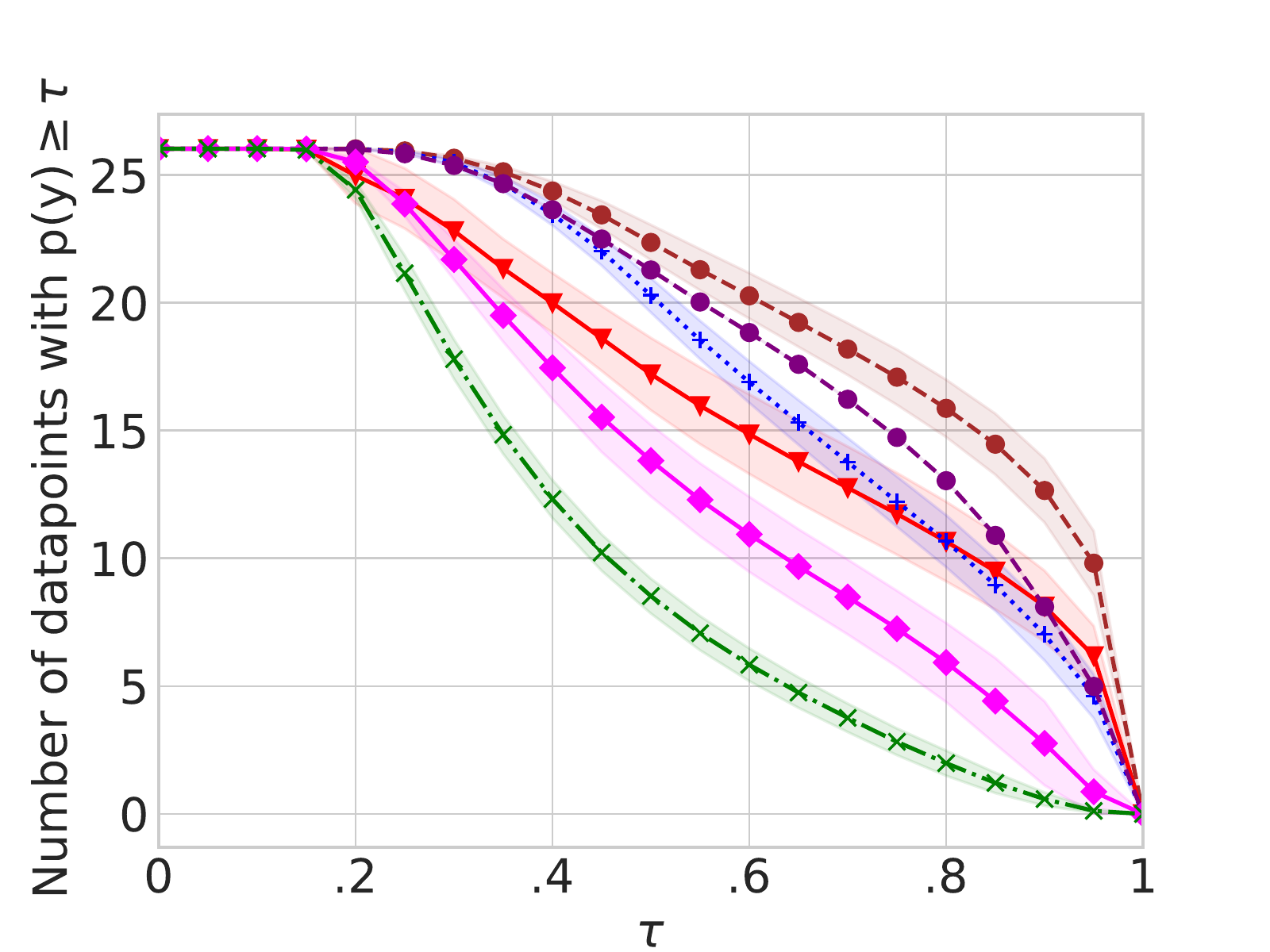}}

\end{center}

\caption{Performance  under varying degrees of rotation in CIFAR10 (top) and SVHN (bottom). LT-NODE and ALT-NODE show better uncertainty modelling capability with higher LL values and lower Brier and ECE scores as we increase the rotation. Confidence distribution plots  on rotated
(45 degrees) images are shown in (d), (e), (i) and (j). (d) and (i) shows
the accuracy, and (e) and (j) shows the count of predictions
(y axis value multiplied by 1000) done with confidence above
a threshold $\tau$ on X-axis. The proposed models perform better
than baselines giving higher accuracy's and lower counts on
high confident predictions.
} 
\vspace{-0.5cm}
\label{fig:rotation}
\end{figure*}
\begin{figure*}[t]
\begin{center}     
\subfigure[Error (CIFAR10)]{\includegraphics[scale=0.2]{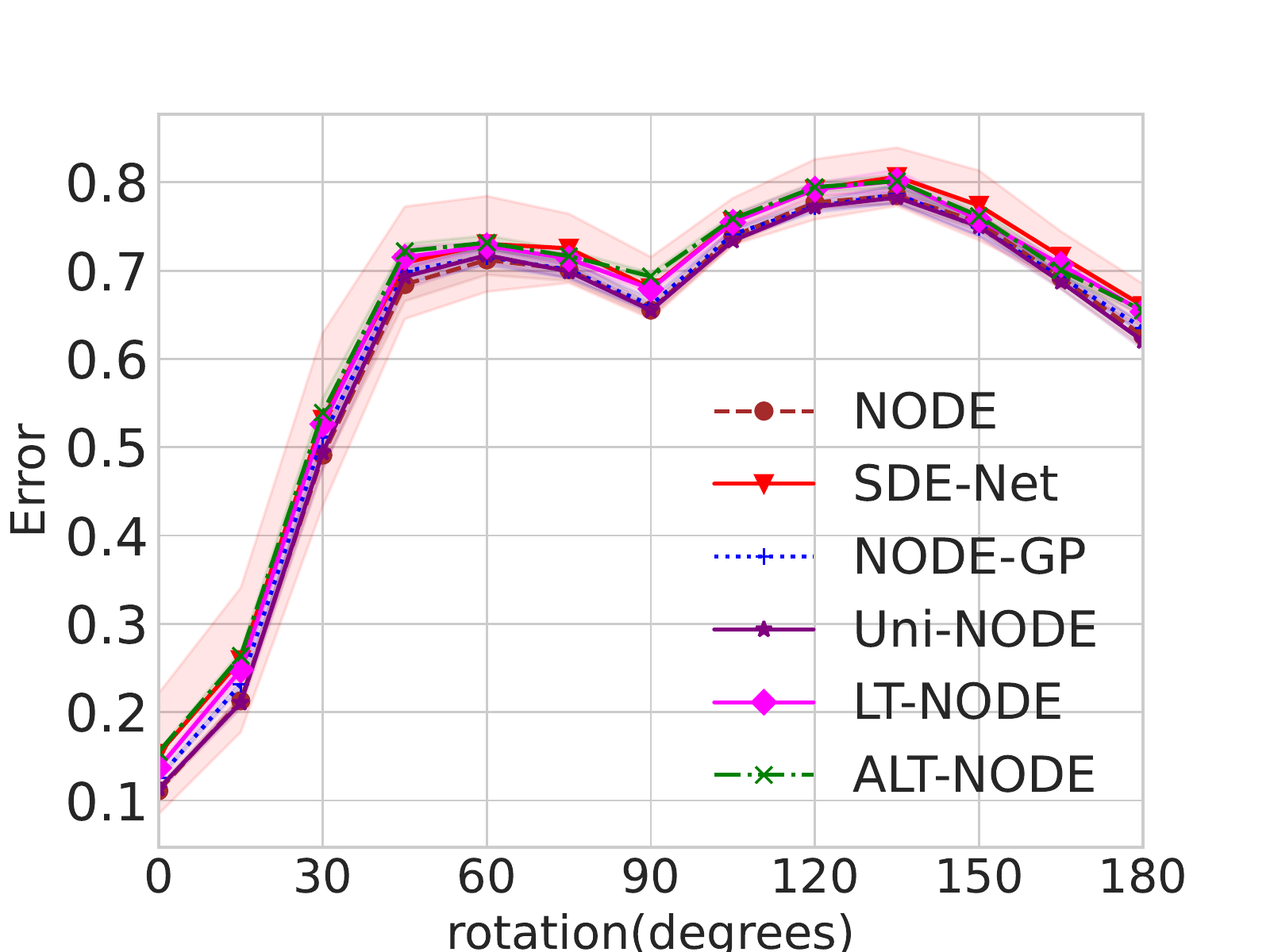}}
\subfigure[CIFAR10-SVHN OOD Rejection]{\includegraphics[scale=0.2]{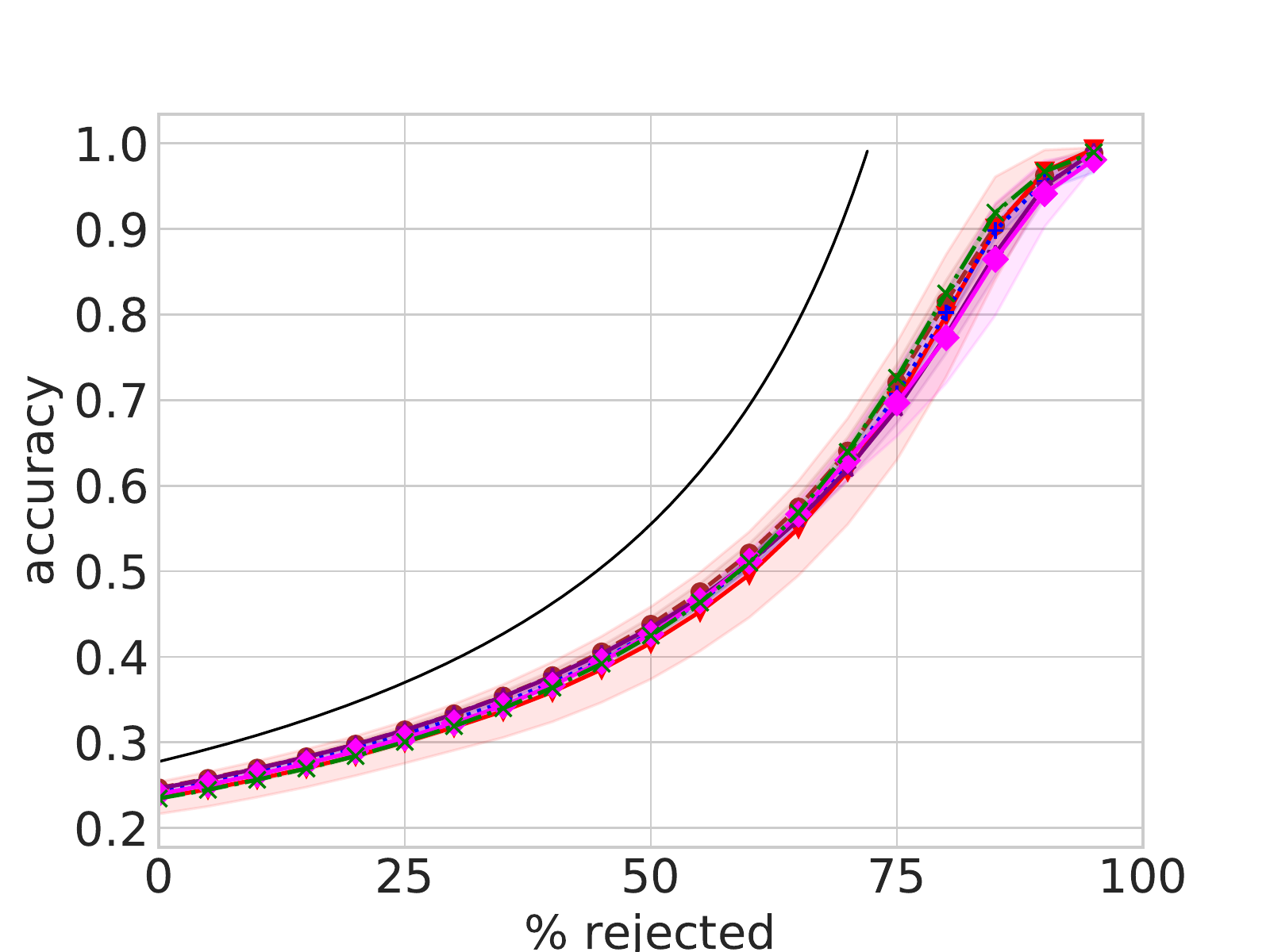}}
\subfigure[CIFAR10-SVHN Entropy Histogram]{\includegraphics[scale=0.2]{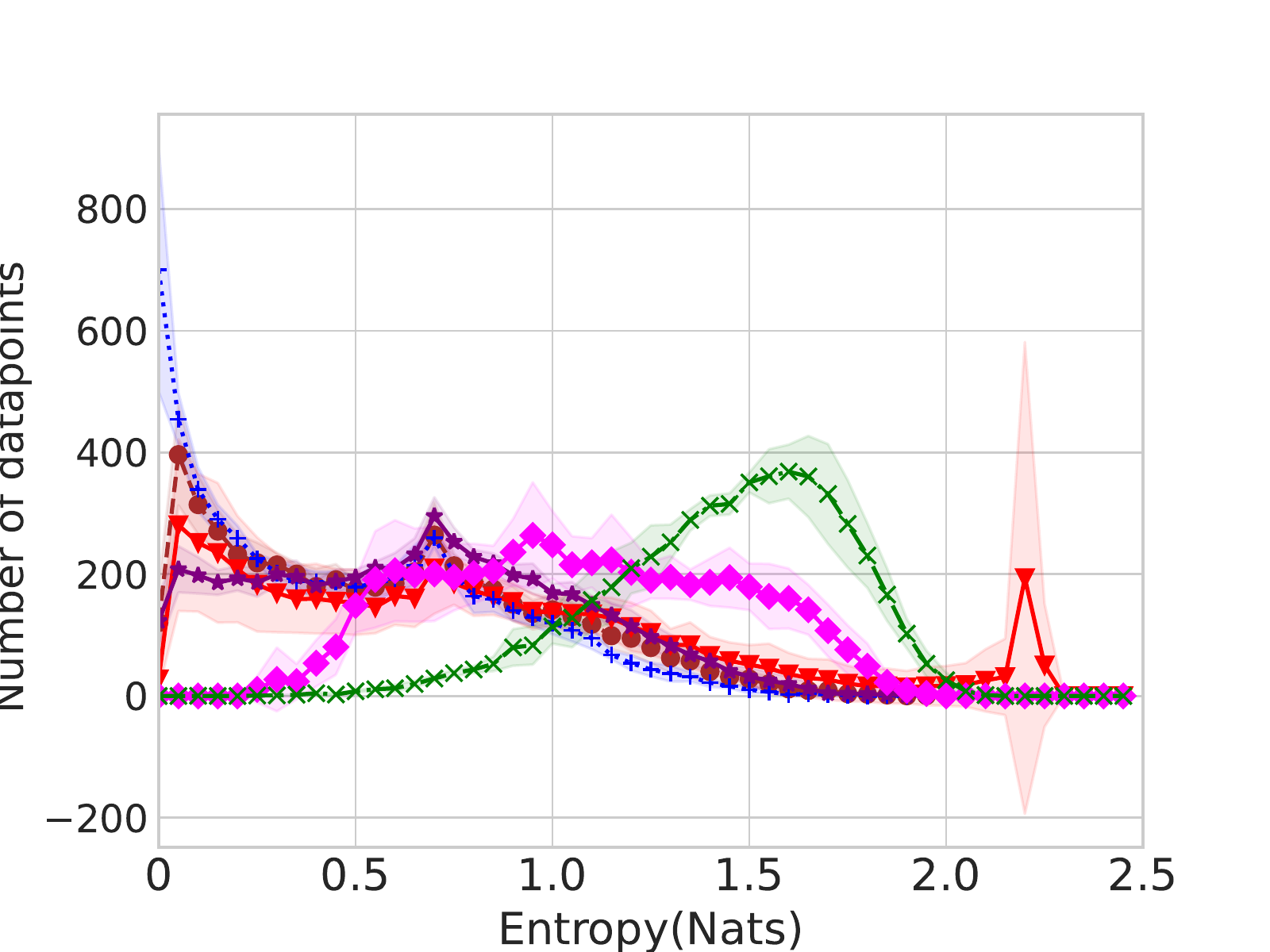}}
\subfigure[MNIST]{\includegraphics[scale=0.2]{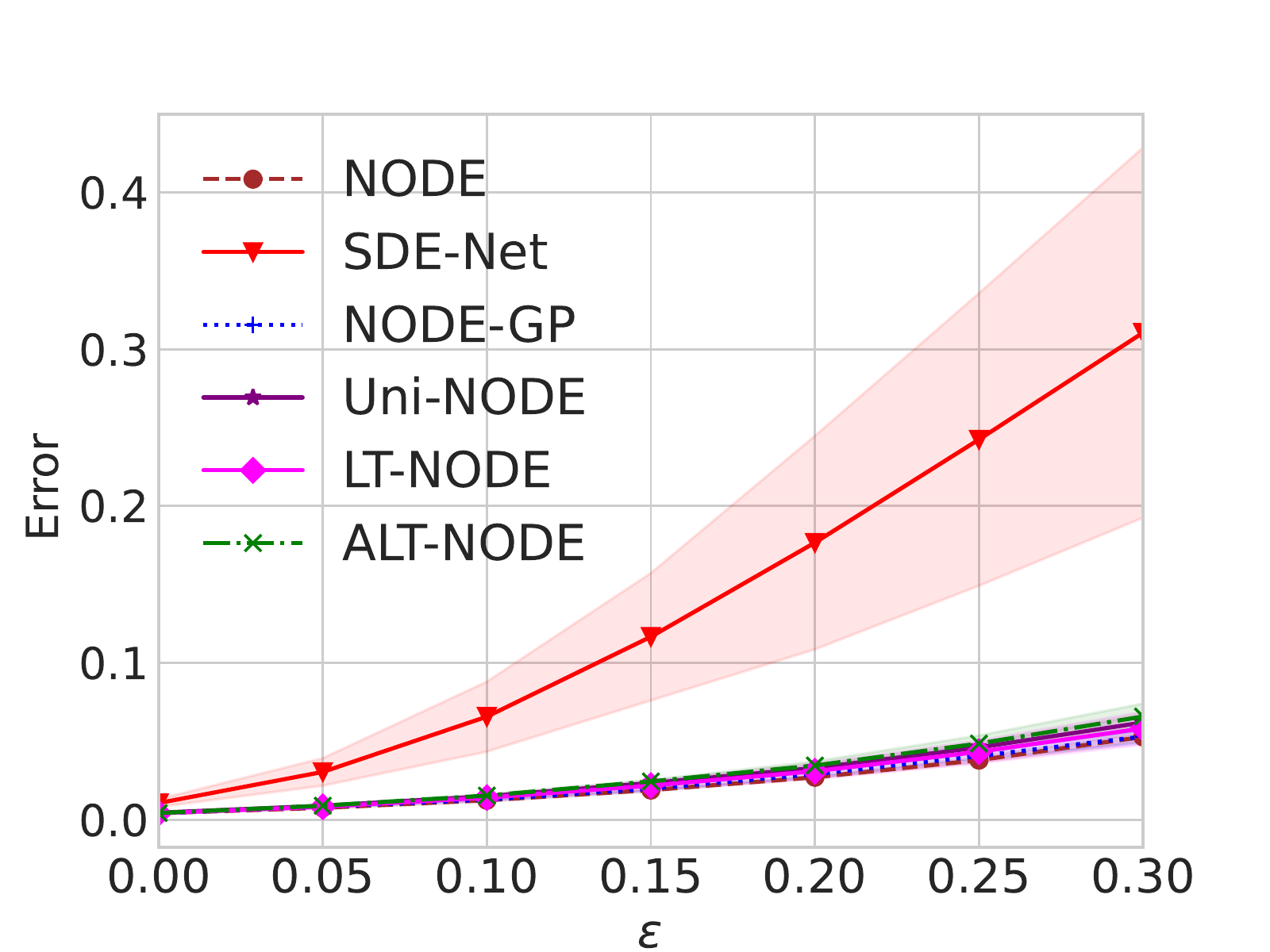}}
\subfigure[F-MNIST]{\includegraphics[scale=0.2]{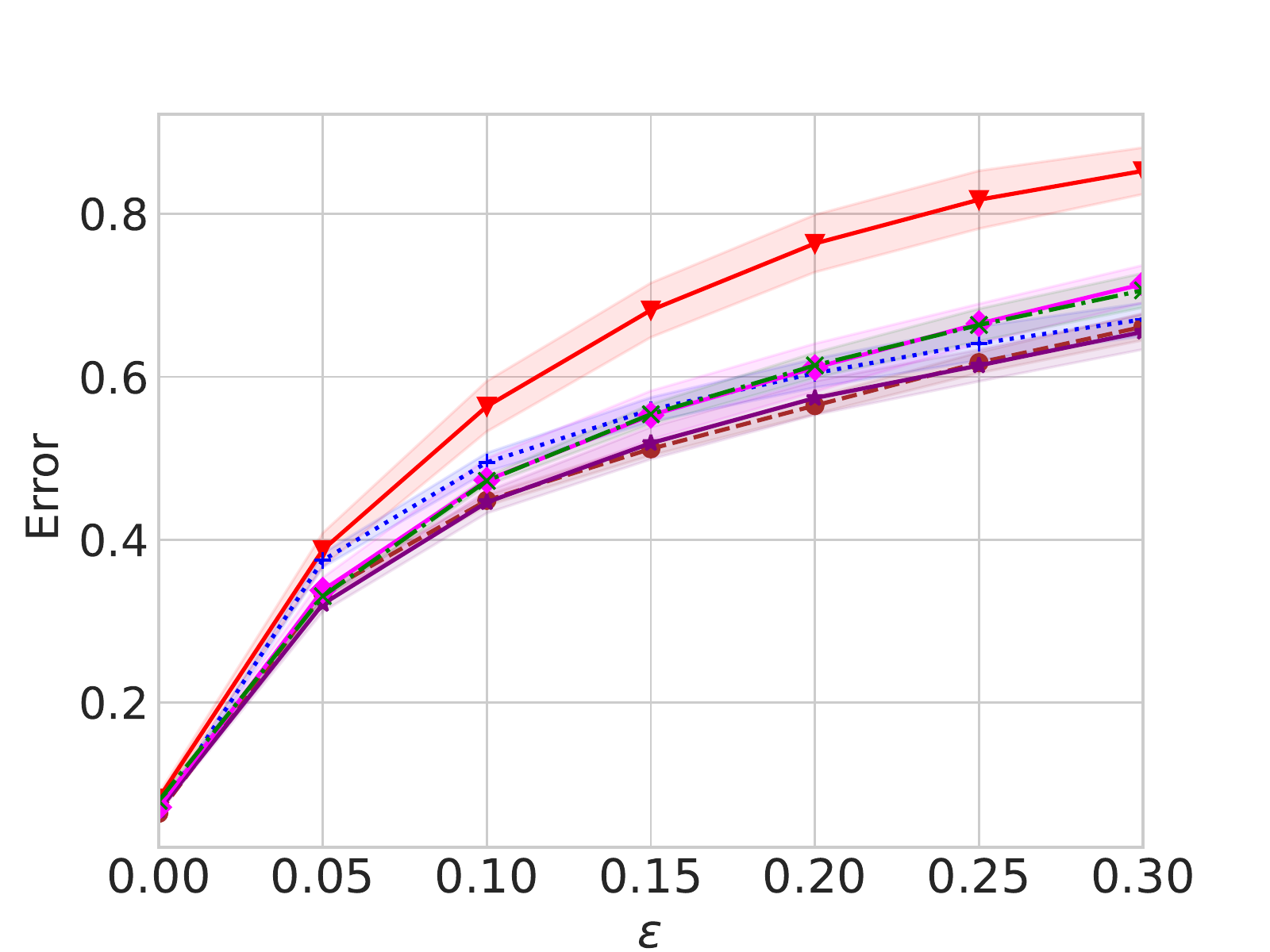}}
\subfigure[Error (SVHN)]{\includegraphics[scale=0.2]{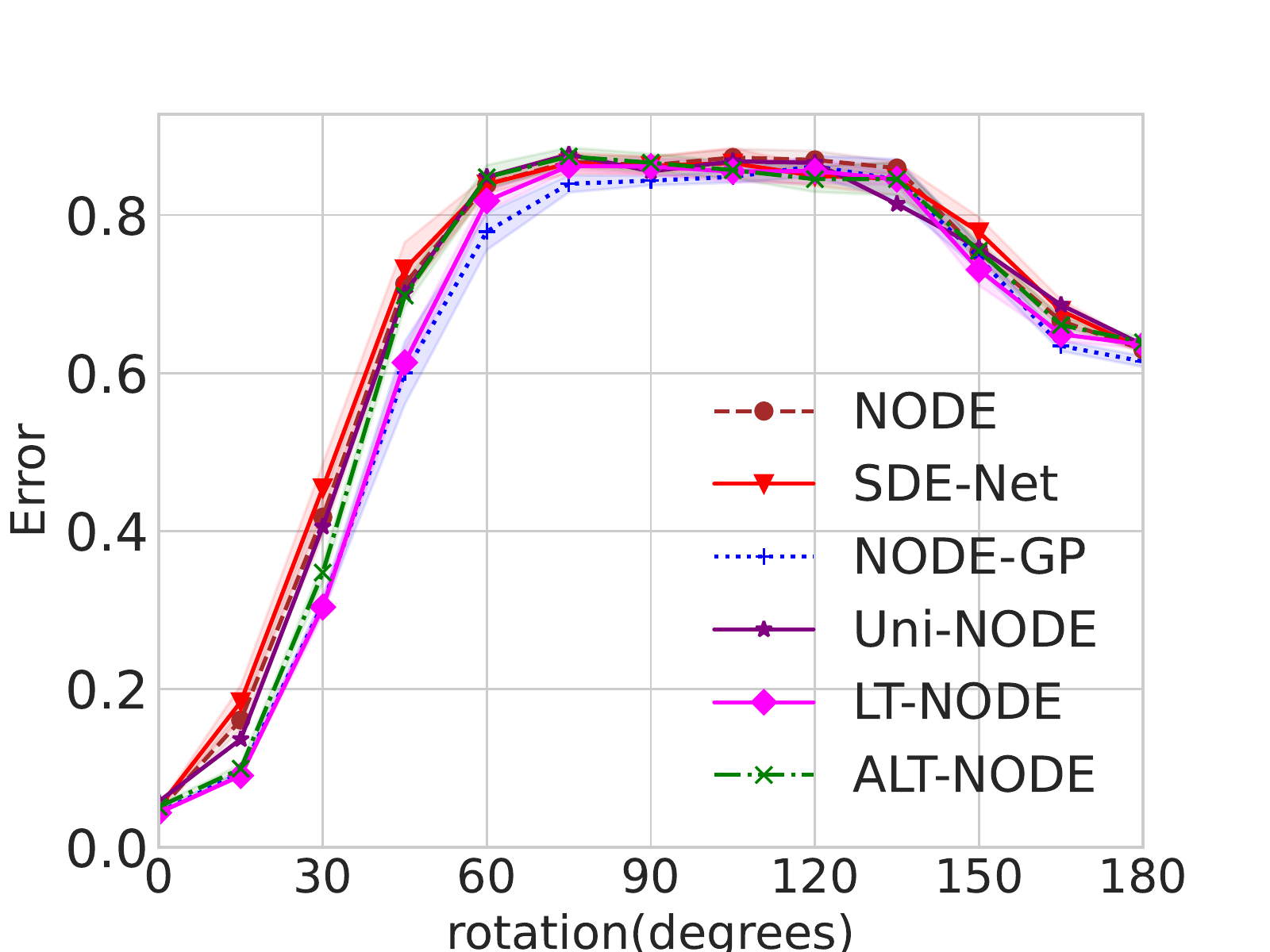}}
\subfigure[SVHN-CIFAR10 OOD Rejection]{\includegraphics[scale=0.2]{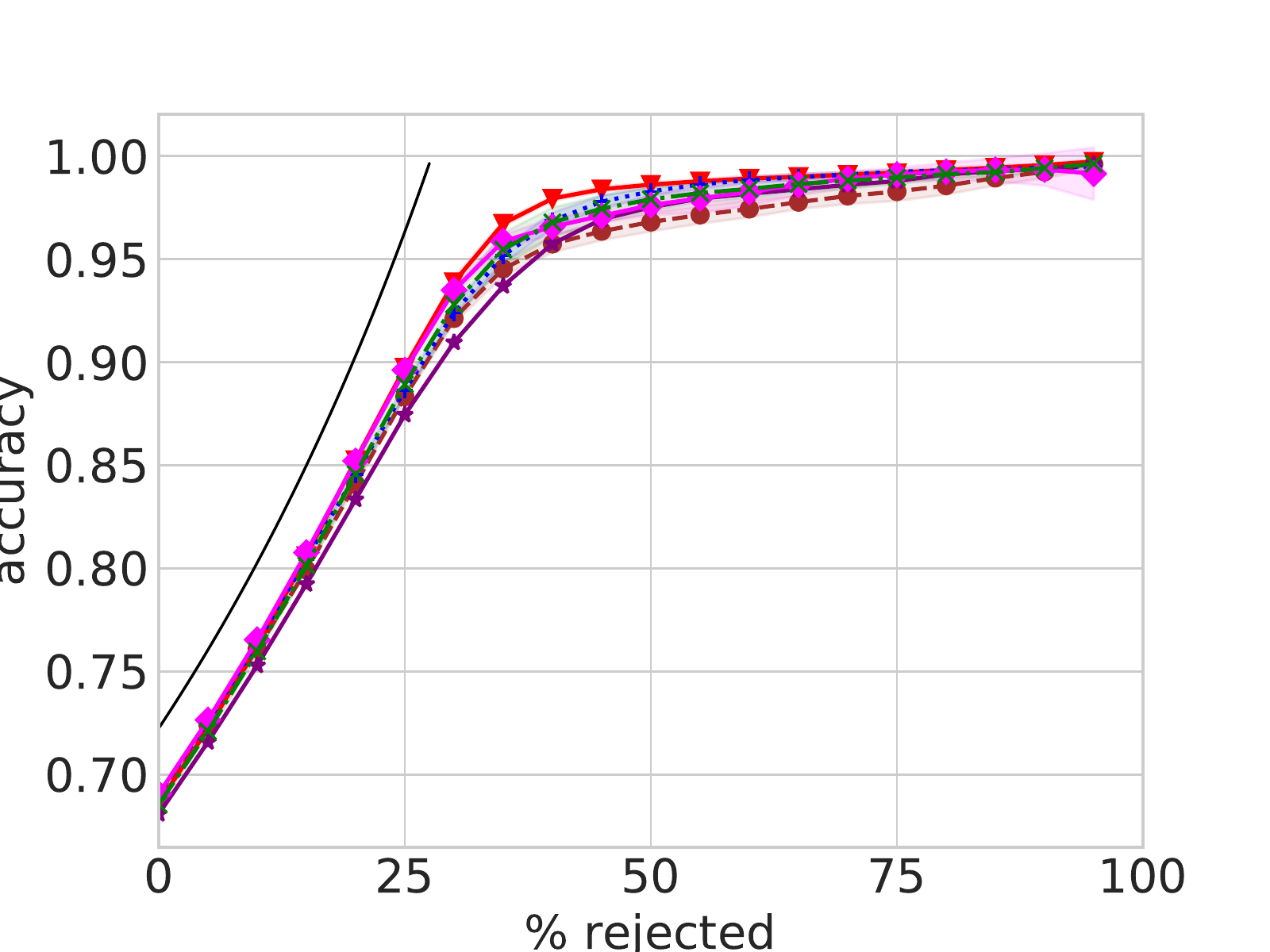}}
\subfigure[SVHN-CIFAR10 Entropy Histogram]{\includegraphics[scale=0.2]{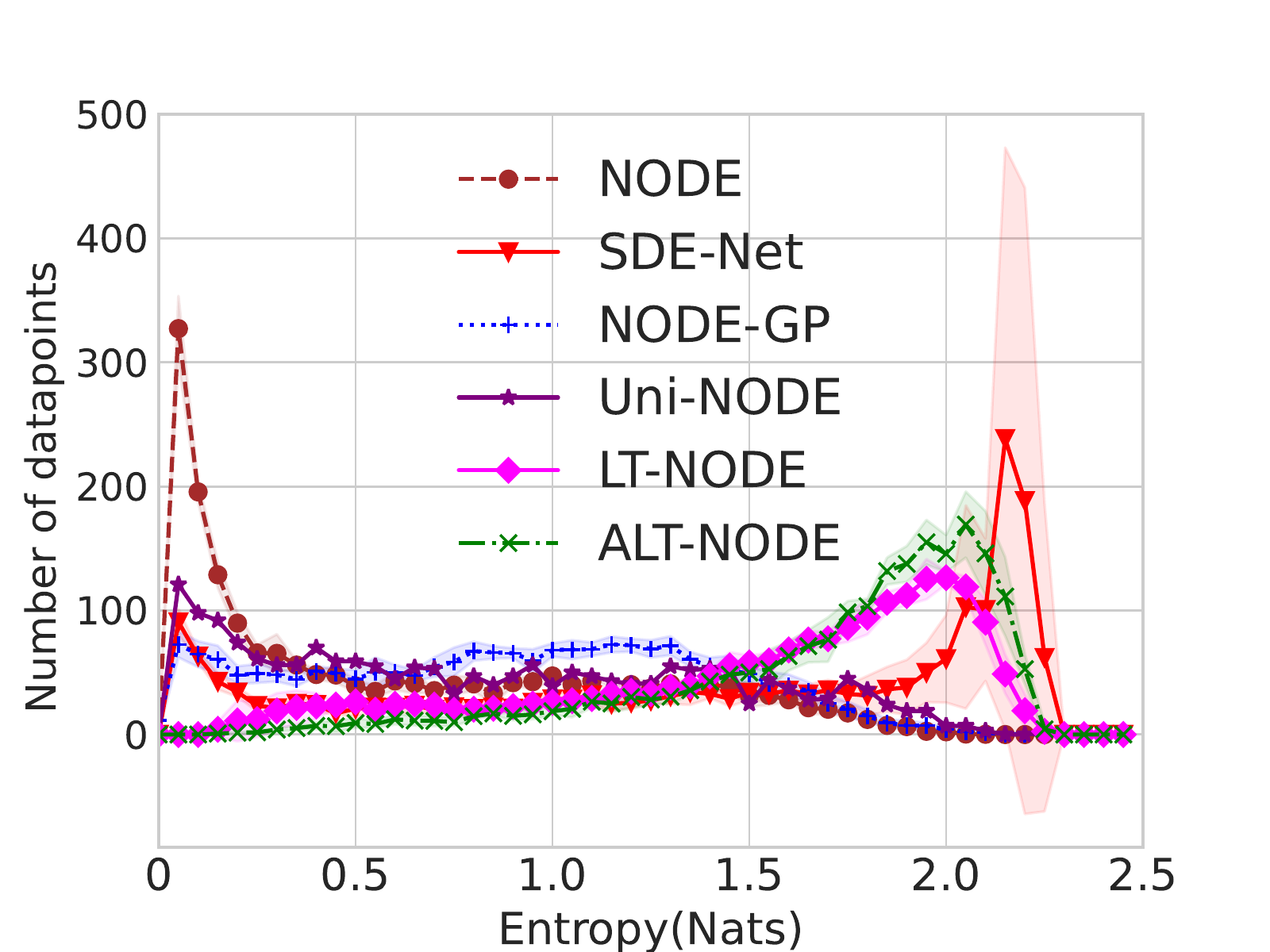}}
\subfigure[CIFAR10]{\includegraphics[scale=0.2]{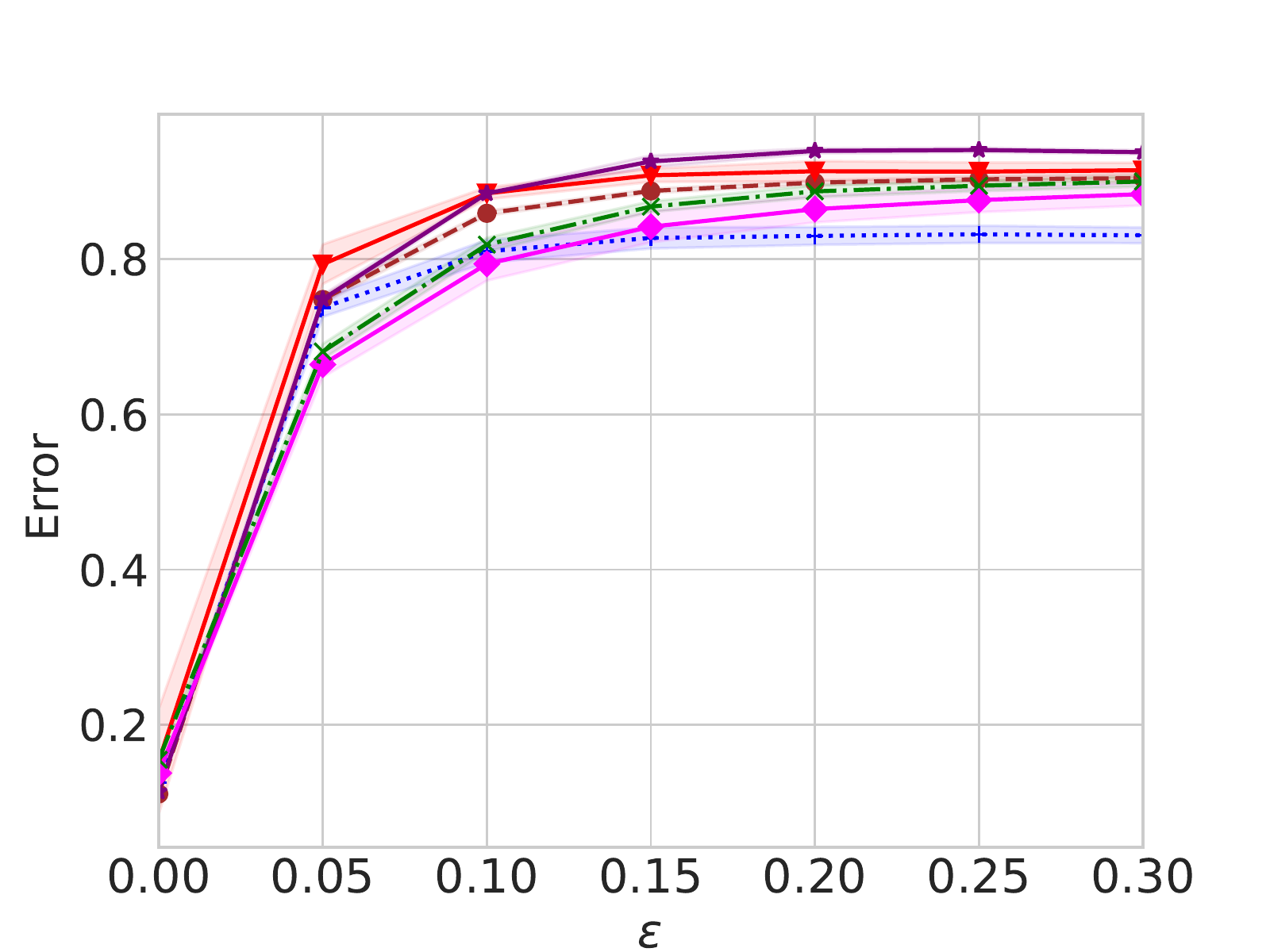}}
\subfigure[SVHN]{\includegraphics[scale=0.2]{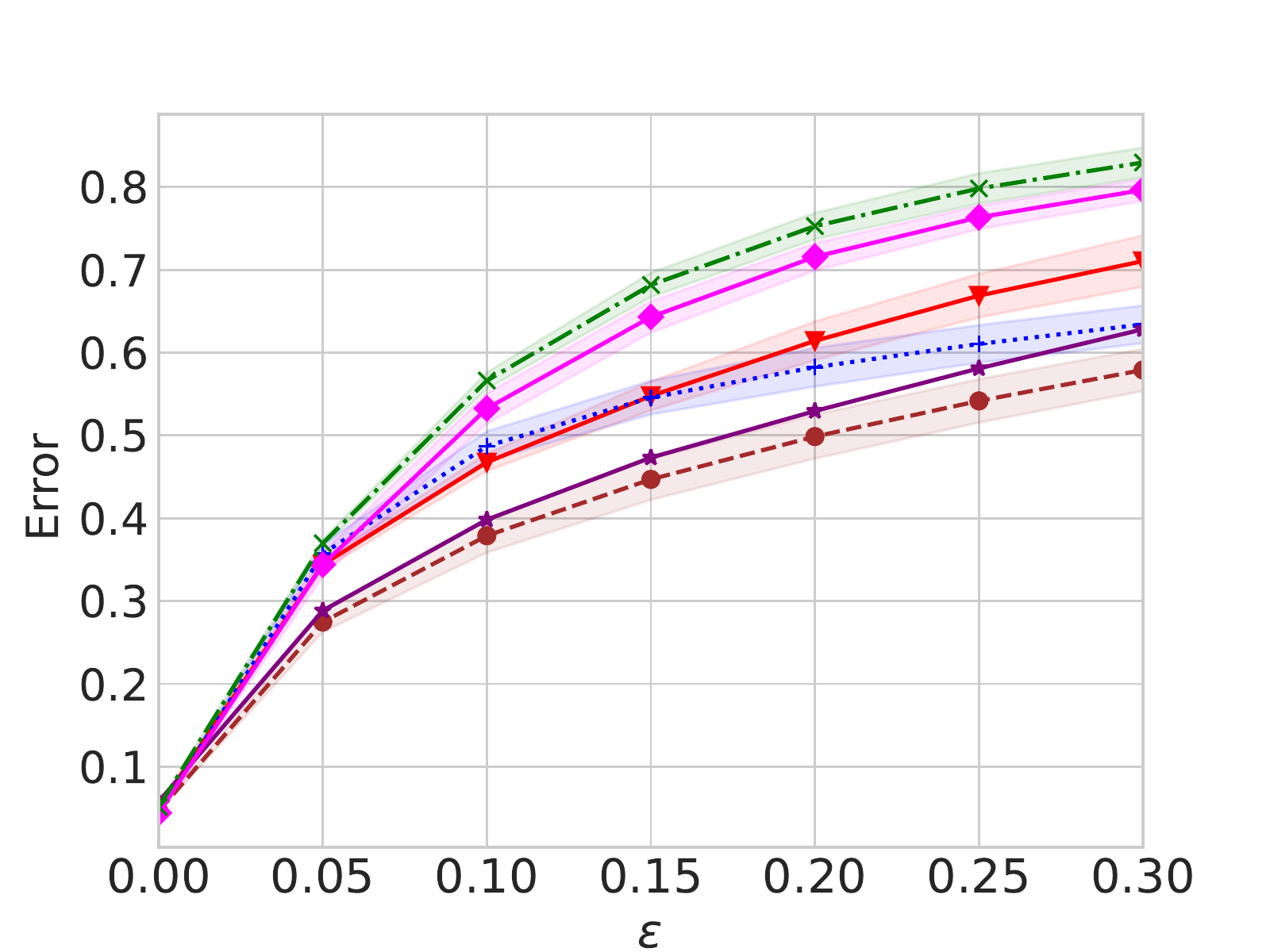}}
\end{center}
\vspace{-1em}
\caption{Error of the models under varying degrees of
rotation in CIFAR10 (a) and SVHN (f). OOD Rejection plot (b and g) and Entropy histogram (c and h) of the models for  OOD experiments. For (b) and (g), training is done on CIFAR10 and testing on SVHN, while for (c) and (h) training is done on SVHN and testing on CIFAR10. Error obtained by the models under FGSM attack
against varying $\epsilon$ (stepsize) on various data sets (d) MNIST (e) Fashion-MNIST (i) CIFAR10 and (j) SVHN.  } 
\label{fig:OOD}
\vspace{-1em}
\end{figure*}
\subsection{Image classification}
We conduct experiments to study uncertainty modelling and robustness capability of the proposed models  on image classification problems. We consider popular data sets used in image classification such as CIFAR10~\cite{krizhevsky2009learning}, SVHN~\cite{netzer2011reading}, MNIST~\cite{lecun1998gradient} and Fashion-MNIST~\cite{xiaofashion}. 
To measure their uncertainty modelling capability, we use several metrics such as Error, log-likelihood (LL), Bier score  and expected calibration error (ECE). Error ($1-$ accuracy) and LL are the standard metrics used in image classification.
LL consider the probability distribution over outputs and can measure the uncertainty modelling capability of the models (higher the better).  
Brier score~\citep{brier} and ECE~\citep{naeini2015obtaining}  are calibration metrics 
which tells us if the predictive  probability  of the model for a class label  is close to the true proportion of those classes in the  test data. 
Both the measures consider predictive probability and consequently can be used to measure uncertainty in predictions (lower values of Brier score and ECE are preferred). All the models follow the same architecture as standard NODE\footnote{https://github.com/rtqichen/torchdiffeq}.  Additional networks are  required for SDE-Net for diffusion and  ALT-NODE for inference, both using 3 convolution layers followed by a fully connected layer~\footnote{Details of experiments and architecture is in the supplementary. }.
The  plots show mean and standard deviation  obtained by training the models with 5-different initializations.
 
\subsubsection{Performance under rotation} We use CIFAR10 and SVHN data sets for studying the performance of the proposed models under rotation of images~\cite{ovadia19}. 
The performance of the models degrades quickly with increase in amount of rotation applied on the image test data and can be seen in Figure~\ref{fig:OOD}(a) and (f). We want our models to be least overconfident when there is a significant shift in the data. To study this, in Figure~\ref{fig:rotation} we plot the performance of the models in terms of  LL, Brier score and ECE.
LT-NODE and ALT-NODE  have better Brier score, LL and ECE values compared to the baselines, demonstrating their improved uncertainty modelling capability. 
We consider confidence distribution  for the methods when prediction is done on SVHN and CIFAR10 rotated by 45 degrees.  We plot the accuracy (Figure~\ref{fig:rotation} (d) and (i)) and count of test  data points (Figure~\ref{fig:rotation} (e) and (j)) when predictions are done with confidence above a threshold $\tau$.
In general, we want the counts of points predicted with high confidence to be lower and accuracy to be higher as we are dealing with a  corrupted data. We can observe that this is the  case with the proposed approaches, beating baselines in both rotated CIFAR10 and SVHN. Accuracy of the proposed models are the highest with high confident predictions, making them more reliable. 
We have found that ALT-NODE performance is better than all the models. Learning input specific distribution helps in  capturing uncertainty better. LT-NODE   performed better than all except ALT-NODE, showing that learning a distribution over end-times in general improves uncertainty modeling capability. 
\subsubsection{Performance on Out of Distribution Data}
We conduct experiments to study the performance of the models on out-of-distribution (OOD) data  by training them on either CIFAR10 or SVHN and testing them on the other. We expect a good model to exhibit a high uncertainty on the test data set which is measured using the entropy score. Entropy measures the  spread of the predictive probability across the classes and expects a higher entropy (higher spread) on OOD data. 
We follow the OOD rejection setup~\citep{filos2019systematic}, where we receive data from both in-distribution (ID) and OOD. An ideal model should be able to classify with high accuracy by ignoring the data points with high uncertainty. 
We study the classification accuracy of the models as we increase the proportion of rejected points  in Figure~\ref{fig:OOD}(b) and (g). 
We can observe that the performances of all the models are close. So, we analyse the entropy histogram of the models for the OOD data in Figure~\ref{fig:OOD}(c) and (h). We can observe that for the case where models trained on CIFAR10 and tested on SVHN (Figure~\ref{fig:OOD} (c)), the proposed models have a better entropy histogram.  They have higher number of points with high entropy and vice-versa compared to the baselines, reflecting their superior uncertainty modelling capability on OOD data. In summary, the average entropy values on OOD data (SVHN) when trained on CIFAR10 are, NODE: $0.572 \pm 0.062$, SDE-Net: $0.792\pm0.354$, NODE-GP: $0.495 \pm 0.0298$, Uni-NODE: $0.668 \pm 0.040$, LT-NODE: $1.074 \pm 0.078$, ALT-NODE: $1.444 \pm 0.050$. Our proposed models having the higher entropy values, exhibiting higher uncertainty over OOD data. 
In Figure~\ref{fig:OOD} (h) Although SDE-Net having larger number of points on the ends of the histogram spectrum, our proposed models have the better entropy values. NODE: $0.617 \pm 0.021$, SDE-Net: $1.421\pm0.078$, NODE-GP: $0.885 \pm 0.034$, Uni-NODE: $0.808 \pm 0.111$, LT-NODE: $1.537 \pm 0.067$, ALT-NODE: $1.723 \pm 0.034$.
\subsubsection{Robustness Evaluation} 
Deep learning models are prone to adversarial attacks~\citep{goodfellow2016deep,szegedy2013intriguing}. To check robustness of models, we conduct experiments to evaluate their performance under FGSM~\citep{FGSM} attack on  MNIST, F-MNIST, CIFAR10 and SVHN. Figures~\ref{fig:OOD}(d),(e),(i), and (j) shows the robustness of the models in terms of error against  increasing perturbation strength $\epsilon$ of FGSM attack. The proposed models LT-NODE and ALT-NODE performed well with low error  on all the data sets except SVHN, demonstrating their robustness against adversarial attack.
\section{Conclusion}
We proposed a novel method to model uncertainty in NODE by learning a distribution over latent end-times. The proposed approaches can compute uncertainty efficiently in a  single forward pass and helps in end-time selection in NODE. The proposed models, LT-NODE and ALT-NODE were shown to have good uncertainty modelling and robustness capabilities through experiments on synthetic and real-world data image classification data.  We expect to further improve their performance  by considering a multi-modal variational posterior distribution  as a future work.  The proposed NODE models could bring advances in   computer vision applications like autonomous driving  where uncertainty modeling is important.

\section{Acknowledgements}
We acknowledge the funding support from MoE, Govt of India and DST ICPS and computational support through JICA funds. 

\bibliography{aaai22}

\clearpage


\section{Appendix }

We provide more details about the latent time neural ODE (LT-NODE) and its variant, adaptive latent time neural ODE (ALT-NODE) in this supplementary. To summarize, LT-NODE and ALT-NODE treats end-time $T$ in the ODE solver as a latent variable, puts a distribution over $T$ and learns a posterior from the data  using Bayes Theorem. ALT-NODE additionally assumes the end-time to be different for different inputs. Due to intractability, LT-NODE and ALT-NODE uses variational inference and amortized variational inference respectively.  As already shown in the main paper, LT-NODE and ALT-NODE can provide a very \textbf{good uncertainty modelling} capability which surpasses the state-of-the-art NODE networks which can model uncertainty. Along with this, it can help in \textbf{model selection} (selection of end-Time $T$ ) and provides a  very \textbf{ efficient } uncertainty prediction. The proposed approach only requires a \textbf{single forward pass} through the model to compute predictive probability, while prior works required multiple forward passes. LT-NODE besides models uncertainty with just \textbf{2 additional number of parameters} (variational parameters of the gamma posterior).  In this appendix, we provide a detailed derivation of the ALT-NODE and other experimental results demonstrating efficacy of  the proposed models.
\section{Variational lower bound derivation for ALT-NODE}

let $\mathcal{D} =  \{X,\by\} = \{(\bx_i,y_i)\}_{i=1}^{N}$ be the set of training data points with input $\bx_i \in \cR^D$ and $y_i \in \{1,\ldots,C\}$  for a classification and $y_i \in \cR$ for regression. ALT-NODE assumes that every data point to be  associated with a separate latent end-time $T_i$. We assume a Gamma prior over $T_i$ with parameters $\alpha_p$ and $\beta_p$ and the  likelihood being $p(y_i|T,\bx_i,\pmb{\theta})$. We assume an approximate variational Gamma posterior over $T_i$,  $q(T_i| \bx_i, \phi)$ parameterized by the inference network  $r(\bx_i; \phi)$  which predicts the  parameters $\alpha_{qi}$ and $\beta_{qi}$ of the Gamma distribution.
$$
p(T_i|y_i,\bx_i;\pmb{\theta}) \approx  \text{Gamma}(\alpha_{qi},\beta_{qi})  = q(T_i| \bx_i, \phi),
$$
$$
 \quad \text{where} \quad (\alpha_{qi},\beta_{qi}) = r(\bx_i; \phi)
$$

To derive  the variational lower bound for  ALT-NODE,  we start with the evidence of marginal likelihood for a data point $\bx_i$ 
$$p(y_i|\bx_i;\pmb{\theta}) = \int p(y_i|T_i,\bx_i;\pmb{\theta})p(T_i|\alpha_p,\beta_p) dT_i.$$
The evidence over the entire data is given by 
\begin{multline}
        \hspace{-0.5cm}
         \log \prod_{i=1}^{N}p(y_i|\bx_i;\pmb{\theta}) = \log \prod_{i=1}^{N}  \int p(y_i|T_i,\bx_i;\pmb{\theta})|p(T_i|\alpha_p,\beta_p)dT_i \nonumber
\end{multline}
We now introduce the variational distribution $q(T_i| \bx_i, \phi)$ to the evidence 
\begin{multline}
   \log \prod_{i=1}^{N}p(y_i|\bx_i;\pmb{\theta})   \\ =\log\prod_{i=1}^{N}  \int p(y_i|T_i,\bx_i;\pmb{\theta})|p(T_i|\alpha_p,\beta_p)\frac{q(T_i|\bx_i,\phi)}{q(T_i|\bx_i,\phi)}dT_i \\
        = \sum_{i=1}^{N} \log   \int p(y_i|T_i,\bx_i;\pmb{\theta})|p(T_i|\alpha_p,\beta_p)\frac{q(T_i|\bx_i,\phi)}{q(T_i|\bx_i,\phi)}dT_i 
\end{multline}
Applying Jensen's inequality ($f(\mathbb{E}[\bx]) \geq \mathbb{E} [f(\bx)]$ for concave function $f$) and since $\log$ is a concave function, we can get the evidence lower bound as
\begin{eqnarray}
& & \geq \sum_{i=1}^{N} \int q(T_i|\bx_i,\phi) \log  \frac{ p(y_i|T_i,\bx_i;\pmb{\theta})|p(T_i|\alpha_p,\beta_p)} {q(T_i|\bx_i,\phi)} dT_i \nonumber \\
& & = \sum_{i=1}^{N} \biggl [ \mathbb{E}_{q(T_i|\bx_i, \phi)}[\log(p(y_i|T_i,\bx_i;\pmb{\theta}))] \nonumber \\
& & -\mathbb{KL}((q(T_i|\bx_i, \phi)||p(T_i|\alpha_p,\beta_p))) \biggl ].
\end{eqnarray}
As we can observe from the  lower bound, the inference network parameters $\phi$ are shared across the examples allowing statistical strength to be shared and prevents  the number of variational parameters to grow with the data. Inference network also allows us to compute the approximate posterior distribution for new test data point.

\section{Experimental Setup}

\subsection{Evaluating Uncertainty Estimates}

We use metrics such as Error, log-likelihood (LL), Bier score  and expected calibration error (ECE) to measure uncertainty. Error is $1-$ accuracy, and log likelihood is log probability of predicting the correct class label. 
LL can measure the uncertainty modelling capability  and higher value of the LL is better.  
Brier score~\citep{brier} and ECE~\citep{naeini2015obtaining}  are specifically designed to measure the calibration and uncertainty modelling capabilities of the models.  
ECE~\citep{naeini2015obtaining} consider dividing the confidence range $[0,1]$ into different intervals and forming bins consisting of examples with highest predictive probability in some interval. ECE is computed as a weighted average of the absolute difference between average predictive confidence of points in a bin and empirical accuracy of the bin (lower value of ECE is better). Brier score~\citep{brier}  is computed as the squared error between predicted probability of a class and one hot vector representation of the class (as per ground truth), averaged across all classes and examples (Lower value of Brier score is better).

\subsection{Synthetic Dataset Experiments} To demonstrate the uncertainty modeling capability of the models, we considered 1D synthetic regression dataset~\citep{foongbetween}. This dataset contains two disjoint clusters of around 1500 training points.  
We want the models to express high uncertainty in-between the clusters and also away the training data.

\subsubsection{Experimental details} 
Architecture for all the models can be broken down as Input block, NODE block and Output block. Except for  NODE-GP, for all the models both Input block and NODE block each contain 4 fully connected layers, with [50,100,150,50] neurons in Input block and [100,150,100,50]  neurons in NODE block. For NODE-GP, Input block has 1 fully connected layer with neurons [50] , NODE block has 2 fully connected layers with neurons [100,50].   For all the models, output block transforms the feature vector of dimension 50 to 1. For learning model parameters, for all the models except for NODE-GP we set, learning rate as 0.001, momentum as 0.9, weight decay as $1 \times 10^{-4}$, optimizer as SGD,  loss function as mean square error and number of iterations as 3000. For NODE-GP, we used Adam  optimizer with learning rate 0.01, weight decay 0,  mean square error as loss function and number of iterations as 600. 
We found these settings to give good results for the NODE-GP model. For LT-NODE,  learning rate for variational parameters is 0.001 with momentum 0.9 and weight decay 0.  For ALT-NODE model, the learning rate for the inference network $r(\bx_i; \phi)$ is $0.001$, weight decay $5 \times 10^{-4}$, momentum 0.9. Prior is typically a  Gamma distribution with values  $\alpha_p = 2.0$ and $\beta_p = 0.5$  as shape and rate parameter values. 
During training, for computing the approximate expectation end-Time $T_i$s are sampled from a uniform  distribution with boundary values $0$ and $3$. During prediction, $T_i$s are sampled from the learnt posterior $q(T|\alpha_q,\beta_q)$. For Uni-NODE model during prediction $T_i$s are sampled from Uniform distribution with boundary points $0$ and $3$. 
Learning rate schedule is used for both LT-NODE and ALT-NODE models, which is decayed by a factor of 10 at every 1000 iterations. We used Dopri5, an adaptive numerical method as a solver, with $atol$ and $rtol$ set to $10^{-2}$. For SDE-Net using Euler-Maryumma method we tried with different set of values for step size (0.1,0.5) and with different values as number of steps (2,6,10). Number of steps as 1 and step size as 1 worked best for SDE-Net.   For all the models entire dataset of samples 1500 is used a batch for training. SDE-Net uses 10 forward passes for making predictions, while for other models single forward pass will suffice. LT-NODE and ALT-NODE uses 10 samples of end-times and consequently representations, and GP-NODE also consider 10 samples from the final layer GP transformation. 

We conducted all our experiments in PyTorch~\citep{paszke2019pytorch}, a deep learning library, NODE-GP model is implemented with GPyTorch~\citep{gardner2018gpytorch} a Gaussian process inference tool.  For SDE-Net~\citep{kong2020sdenet} we used their available code\footnote{https://github.com/Lingkai-Kong/SDE-Net}. For NODE-GP, we used the available code\footnote{https://github.com/srinivas-quan/NODE-GP} which is built on Gpytorch~\citep{gardner2018gpytorch}\footnote{https://docs.gpytorch.ai/en/v1.1.1/examples/ \\ 06\_PyTorch\_NN\_Integration\_DKL/Deep\_Kernel\_Learning \\ \_DenseNet\_CIFAR\_Tutorial.html}.

\subsubsection{ Approximate  Posterior and Model Selection }
\begin{figure*}[t]

\begin{center}     
\subfigure[Prior ]{\includegraphics[scale=0.5]{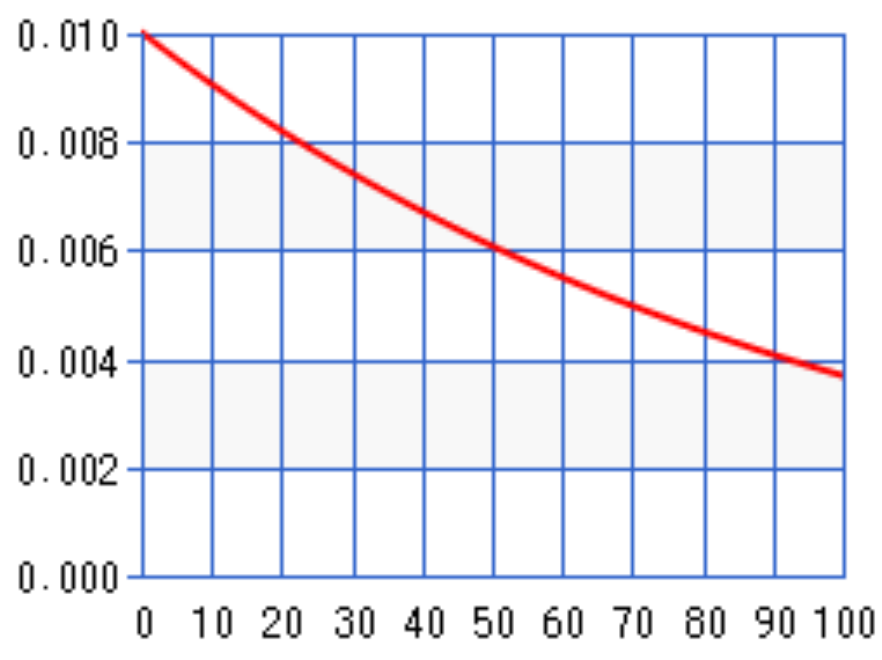}}
\subfigure[Variational Posterior]{\includegraphics[scale=0.5]{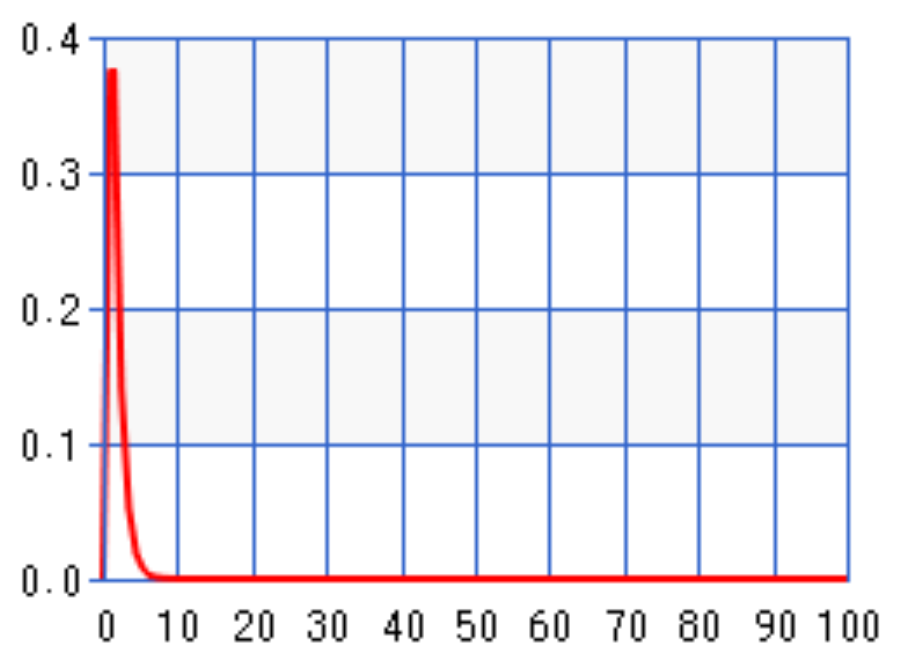}}

\end{center}

\caption{(a) Gamma as prior(1,0.01) distribution for training synthetic data using LT-NODE (b) Learnt posterior distribution(1.05, 0.99) which is a Gamma distribution 
} 
\label{fig:synthetic1}

\end{figure*}

\begin{figure*}[t]

\begin{center}     
\subfigure[Prior ]{\includegraphics[scale=0.5]{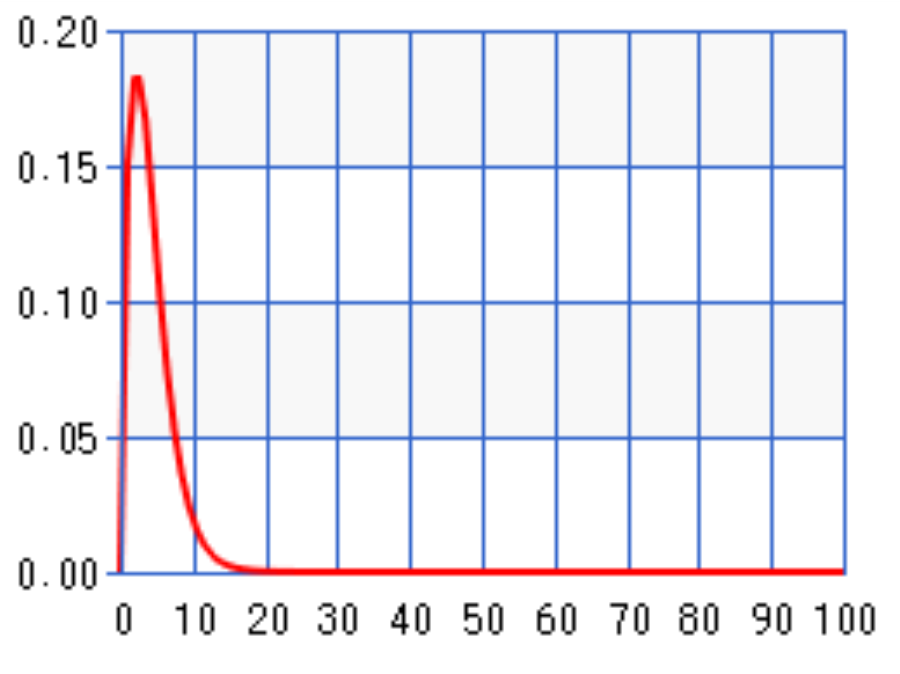}}
\subfigure[Variational Posterior]{\includegraphics[scale=0.5]{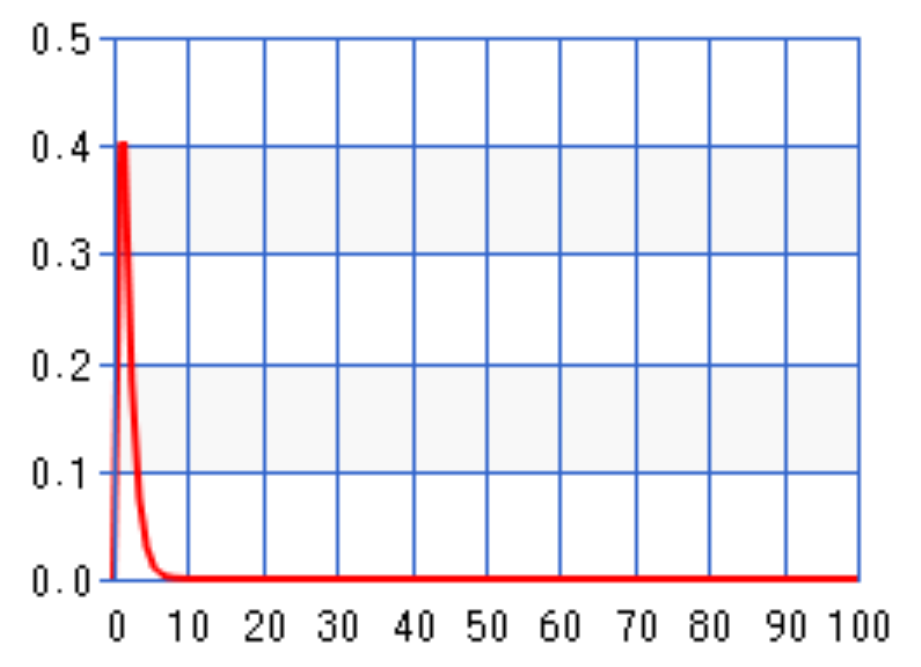}}

\end{center}

\caption{(a) Gamma as prior(2,0.5) distribution for training synthetic data using LT-NODE (b) Learnt posterior distribution(1.27,0.98) which is a Gamma distribution 
} 
\label{fig:synthetic2}

\end{figure*}

We plot the approximate posterior over $T$ learnt using the LTNODE in Figure~\ref{fig:synthetic1} and Figure~\ref{fig:synthetic2}. We conduct experiments by considering different prior  distribution over $T$, such a $\text{Gamma}(1,0.01)$ prior and $\text{Gamma}(2, 0.5)$ prior. LTNODE learnt an approximate posterior  $\text{Gamma}(1.05,0.99)$ and $\text{Gamma}(1.27,0.98)$ respectively. We can observe that the model has learnt a similar posterior in both the cases, with a high density in the lower values of $T$. This  implies that we are learning a correct posterior irrespective of the prior. However, we note that learning of variational parameters from ELBO is not a convex optimization problem and can lead to a  different  posterior. Another interesting observation is that the density is high for low values of $T$, and consequently we sample small values of $T$. This means that the model requires very few transformations to predict correctly. This corroborates well with the extensive model selection we conducted for various models, where we found that lower values of $T$ gives a  good performance. 

\begin{table}
  \caption{Hyperparameter values for training Image datasets}
  \centering
  \begin{tabular}{l|lll}
    \toprule
    Model & learning\_rate & weight\_decay & Momentum \\
    \midrule
    NODE  & $0.1$ & $0.9$ & $5\times10^{-4}$ \\
    SDE-Net & $0.1$ & $0.9$ & $5 \times 10^{-4}$ \\
    NODE-GP & $0.1$ & $0.9$ & $1 \times 10^{-4}$ \\
    Uni-NODE & $0.1$ & $0.9$ & $5 \times 10^{-4}$ \\
    LT-NODE & $0.1$ & $0.9$ & $1 \times 10^{-4}$ \\
    ALT-NODE & $0.1$ & $0.9$ & $1 \times 10^{-4}$ \\
    \bottomrule
  \end{tabular}
  \label{tab:hyper}
\end{table}

\subsection{Image Classification}
Performance of the proposed models is evaluated on real-world datasets such as MNIST~\citep{lecun1998gradient}, F-MNIST~\citep{xiaofashion}, SVHN~\citep{netzer2011reading}, CIFAR10~\citep{krizhevsky2009learning}. Summary about the image datasets can be found in Table \ref{tab:datasets}.

\subsubsection{Training}
All the models are trained as per the configuration provided in Table \ref{tab:config}. Every model on every dataset is trained for 5 times with different model initializations. We followed the configuration~\citep{antoran2020depth} for selecting number of epochs and learning rate schedules to train  the models. All the models are trained using SGD with momentum and weight decay. The initial learning rate for all the models is $0.1$. The learning rate is decayed by a factor of $10$ at scheduled epochs. The scheduled epochs for each dataset is provided in Table\ref{tab:config}. For SDE-Net~\citep{kong2020sdenet} we used their available code\footnote{https://github.com/Lingkai-Kong/SDE-Net}, with the configuration as discussed in Table\ref{tab:config}. For NODE-GP, we used the available code\footnote{https://github.com/srinivas-quan/NODE-GP} which is built on Gpytorch~\citep{gardner2018gpytorch}\footnote{https://docs.gpytorch.ai/en/v1.1.1/examples/06\_PyTorch\_NN \\ \_Integration\_DKL/Deep\_Kernel\_Learning\_DenseNet\_CIFAR\_Tutorial.html}. For LT-NODE model, learning rate for variational parameters is $0.01$ with momentum $0.9$ with weight decay is $0$. For LT-NODE model, while training, end-Time $T_i$s are sampled from a uniform distribution with boundary values $0$ and $3$. During testing, $T_i$s are sampled from the learnt posterior $q(T|\alpha_q,\beta_q)$.  For ALT-NODE model, the learning rate of the inference network $r(\bx_i; \phi)$ is $0.01$, weight decay $5 \times 10^{-4}$, and  momentum 0.9. For both LT-NODE and ALT-NODE the prior is Gamma distribution with values  $\alpha_p = 2.0$ and $\beta_p = 0.5$  as shape and rate parameters.  For all the models batch size is 256. SDE-Net uses 10 forward passes to compute the predictive uncertainty, while for other models single forward pass will suffice. LT-NODE and ALT-NODE uses 10 samples of end-times and consequently representations, and GP-NODE also consider 10 samples from the final layer GP transformation.   All the models are trained on Nvidia P100 GPU. We used the available code\footnote{https://github.com/juntang-zhuang/torch\_ACA} for implementing adaptive numerical method(Dopri5) numerical method. We set $atol$ and $rtol$ values to $10^{-2}.$

\subsection{Inference in LT-NODE} During inference, a set of $S$ samples are drawn from the learnt posterior and are sorted in increasing order. The ODE is solved using a adaptive numerical method(Dopri5) with initial value as $0$ and end-time as $T$(maximum value among all the $S$ samples). While solving the ODE, the accepted time-points and feature vectors are stored in memory(with torch.no\_grad()). The feature vectors at $T_i$s are obtained by interpolating using the accepted times and feature vectors. The computed feature vectors are used for making predictions.

\subsection{Inference in ALT-NODE}

For ALT-NODE, we make the inference efficient through the following procedure. During inference, for the data points in a given batch, we first compute the variational parameters $\alpha_i$ and $\beta_i$ using the inference network. End-times for each input data is sampled from the variational Gamma posterior with parameters $\alpha_i$ and $\beta_i$. The set of sampled end-times for all the data in a batch are sorted in increasing order. The ODE is solved using a adaptive numerical method(Dopri5) with initial value as $0$ and end-time as $T$(maximum value among all the $S$ samples). While solving the ODE, the accepted time-points and state/feature vectors are stored in memory. The state/feature vectors at $T_is$ are obtained by interpolating using the accepted times and feature vectors. The interpolated values are stored in memory. For each data point, the relevant feature vector is extracted from the memory ( according to sampled end-times) and used to compute the output. Note that in this way, we only need to perform a single forward pass through the NODE for all the data points in a batch.

\subsection{Backpropagation in (A)LT-NODE:} While training, the datapoints are feed-forwarded through the downsampling layer. A set of $S$ samples($T_i$s) are drawn from the uniform grid and are sorted in increasing order. The feed-forward in NODE contains two-phases. 

Phase 1: The ODE is solved using a adaptive numerical method(Dopri5) with initial value as $0$ and end-time as $T$(maximum value among all the $S$ samples). During the trajectory computation, computational graph is not constructed(in pytorch this can be achieved using torch.no\_grad()) as it contains the sub branches which are involved in deciding the step-size and consume memory. During the trajectory computation only the accepted times are stored in memory. 

phase 2: Using the accepted times, the trajectory is reconstructed using the dopri5 numerical method. The feature vectors at $S$ samples are computed by interpolating using the accepted times and corresponding feature vectors.  These interpolated feature vectors are feed-forwarded through FCNN layer to make prediction which are further used for computing loss. During backpropagation, the gradient of loss with respect to weights are computed and updated.
\begin{table*}
  \caption{Summary of datasets}
  \centering
  \begin{tabular}{l|llll}
    \toprule
    Dataset     & Size of train data & Size of test data & Input-Dimension & No. Classes \\
    \midrule 
    MNIST & $60,000$ & $10,000$ & $28\times28$ & $10$     \\
    F-MNIST & $60,000$ & $10,000$ & $28\times28$ & $10$ \\
    CIFAR10 & $50,000$ & $10,000$ & $32\times32\times3$ & $10$ \\
    SVHN & $73,257$ & $26,032$ & $32\times32\times3$ & $10$ \\
    \bottomrule
  \end{tabular}
  \label{tab:datasets}
\end{table*}

\begin{table}
    \caption{Training configurations for each dataset}
    \centering
    \begin{tabular}{l|ll}
    \toprule
    Dataset      & Number of epochs & Learning rate schedule \\
     \midrule 
     MNIST       &      90          &      40, 70              \\
     F-MNIST     &      90          &      40, 70              \\
     CIFAR10     &      300          &      150, 225              \\
     SVHN        &      90          &      40, 70              \\
     \bottomrule
    \end{tabular}
    \label{tab:config}
\end{table}

 \begin{table}
     
     \caption{Performance of the models in terms of generalization error. Mean values and standard deviation of the error is given. The column $\#$ parameters are given in Millions. }
     
     \label{tab:model_perform}
     \centering
     \begin{tabular}{l l| l l }
     \toprule
     Dataset & Model &$\#$Parameters & Error \\
      \toprule
         &   NODE   & 0.21M  &$0.005\pm0.000$ \\ 
         &  SDE-Net &0.28M  &$0.010\pm0.002$ \\
   MNIST &  NODE-GP &0.47M &$0.004\pm0.001$\\
         &  Uni-NODE &0.21M  & $0.005\pm0.000$ \\
         &  LT-NODE &0.21M  & $0.004\pm0.001$ \\
         &  ALT-NODE& 0.34M &$0.004\pm0.000$\\
     \bottomrule
           &   NODE & 0.21M  &$0.064\pm0.001$  \\
         &  SDE-Net &0.28M  &$0.082\pm0.006$ \\
   F-MNIST &  NODE-GP &0.47M  &$0.068\pm0.001$ \\
         &  Uni-NODE&0.21M  &$0.064\pm0.001$ \\
         &  LT-NODE &0.21M  &$0.071\pm0.001$ \\
         &  ALT-NODE& 0.34M& $0.078\pm0.002$ \\
     \bottomrule
         &  NODE &0.21M  &$0.111\pm0.002$ \\
         &  SDE-Net &0.32M &$0.153\pm0.068$ \\
   CIFAR10 &  NODE-GP&0.47M &$0.116\pm0.002$ \\
         &  Uni-NODE& 0.21M&$0.114\pm0.003$ \\
         &  LT-NODE &0.21M &$0.131\pm0.002$ \\
         &  ALT-NODE& 0.34M  &$0.155\pm0.002$ \\
     \bottomrule
         &   NODE &0.21M  &$0.046\pm0.001$ \\
         &  SDE-Net &$0.32$M  &$0.050\pm0.001$ \\
   SVHN &  NODE-GP & 0.47M  & $0.044\pm0.001$ \\
         &  Uni-NODE &0.21M  &$0.046\pm0.001$ \\
         &  LT-NODE &0.21M  &$0.044\pm0.001$ \\
         &  ALT-NODE& 0.34M &$0.050\pm0.002$ \\
     \end{tabular}
     \label{tab:generalization}
 \end{table}

 \begin{table*}
     \caption{Out-of-distribution detection results on CIFAR10, SVHN.}
     \label{tab:model_ood}
     \centering
     \begin{tabular}{l l l| l l l }
     \toprule
     Dataset & OOD-Data & Model &AUROC &AUPR In & AUPR Out  \\
      \toprule
         & & NODE & $83.750 \pm 1.629$ & $78.0.35 \pm 2.239$ & $89.625 \pm 1.274$  \\ 
         &  &SDE-Net & $81.272 \pm 5.790$ & $76.497 \pm 5.978$ & $87.466 \pm 4.214$   \\
   CIFAR10 & SVHN & NODE-GP& $82.530 \pm 0.841 $ & $76.415 \pm 1.223 $ & $88.935 \pm 0.613$ \\
         & & Uni-NODE & $82.698 \pm 1.236$ & $75.762 \pm 1.941$ & $89.368 \pm 0.731$ \\
         & &LT-NODE & $83.553 \pm 2.04 $ & $77.524 \pm 2.752$ & $89.383 \pm 1.572$  \\
         &  &ALT-NODE& $84.994 \pm 1.450 $ & $ 80.939\pm 1.067$ & $89.928\pm 1.260$  \\
     \bottomrule
         &   &NODE & $93.048 \pm 0.487$ & $97.006 \pm 0.322$ & $ 82.319 \pm 0.564$  \\
         &  &SDE-Net & $96.91 \pm 0.207$ & $98.719 \pm 0.083$ & $92.394 \pm 0.689$  \\
   SVHN & CIFAR10 & NODE-GP & $ 94.19\pm 0.278$ & $ 97.712\pm 0.134$ & $ 83.825\pm 0.462$   \\
         & &  Uni-NODE  & $ 92.398 \pm 0.954 $ & $96.835 \pm 0.445$ & $81.235\pm 1.437$  \\
         & & LT-NODE & $95.287 \pm 0.291$  & $98.025 \pm 0.205 $ & $88.358 \pm 0.551$  \\
         &  & ALT-NODE & $95.542 \pm 0.695$ & $ 98.178\pm 0.365$  & $ 88.502\pm 1.302$  \\
   \bottomrule
     \end{tabular}
     \label{tab:ood_additional}
 \end{table*}

\subsection{Generalization performance on Image data}

Table \ref{tab:generalization} gives the generalization performance of all the models in terms of mean of the test error over  different initializations. Table \ref{tab:generalization} also provides the number of parameters required for each model with respect to the dataset it is trained. 
LT-NODE and ALT-NODE models provide comparable generalization performance on these image classification data and can in addition model uncertainty better than other models as demonstrated through other experiments. 
However, the  performance of SDE-Net model is generally poor in terms of mean error  and also has the highest standard deviation across all the datasets except for SVHN.  

We also note that the proposed LT-NODE model achieves a good generalization performance and uncertainty modelling capability with almost same number of parameters as NODE (just 2 additional variational parameters ). For LT-NODE the number of learnable parameters are 0.21M, which is equivalent to a simple NODE model. LT-NODE captures the uncertainty well without augmenting any non-parametric model like GP or without the requirement of any extra neural network like diffusion network for SDE-Net. Although, ALT-NODE requires additional parameters to represent the inference network, the network size is similar to diffusion network in SDE-Net. Still ,the performance of ALT-NODE is better than any other model for modeling uncertainty.

\subsection{Out-of-Distribution Detection}
We followed the experimental setup\cite{hendrycks17baseline} for detecting OOD data. In real-world the models will receive data from both in and out-distribution. Uncertainty of the model while making a prediction will help to detect OOD data. We used two metrics for OOD detection. (1) Area under the receiver operating characteristic curve(AUROC) (2) Area under the precision-recall curve(AUPR). Higher values of these metrics indicates that the model is performing better.  Table~\ref{tab:ood_additional} shows the OOD detection performance of all the models trained on the dataset provided in the column dataset and tested on OOD data as given in the column OOD-data. From the Table~\ref{tab:ood_additional}, we can see our proposed models having the highest AUROC, AUPR-In, AUPR-Out values compared to any other models. Among LT-NODE and ALT-NODE, ALT-NODE performs well on OOD-detection experiment. Similarly when the models trained on SVHN and tested on OOD data CIFAR10, our proposed models are comparable with the best performing SDE-Net model in terms of AUROC and AUPR-in.

\subsection{Deployment of FGSM}

Fast Gradient Sign Method(FGSM), it is a one step  white-box adversarial attack, where the adversary has the access to entire model. Given a sample image $\bx_*$, gradient of loss  with respect to each pixel $\nabla_{\bx_*} L (\pmb{\theta},\bx_*,y_*)$ is computed and a proportion of the computed gradient is added to the original image. The hyperparameter required for algorithm is the strength($\epsilon$) of the gradient added to the image. As our data set is normalized, we test the model performance on $\epsilon$ ranging from $0.05$ to $0.3$.
\begin{equation}
    \Tilde{\bx}_* = \bx_* + \epsilon * sign(\nabla_{\bx_*} L (\pmb{\theta},\bx_*,y_*)) 
\end{equation}
 Following the same procedure as discussed in the subsection \textbf{Backpropagation in (A)LT-NODE}, in (A)LT-NODE models the gradient of the loss with respect to each pixel $\nabla_{\bx_*} L (\pmb{\theta},\bx_*,y_*)$ is computed. The  datapoint is added with perturbation noise $\epsilon * sign(\nabla_{\bx_*} L (\pmb{\theta},\bx_*,y_*)) $ such that the perturbed data will maximize the loss($L$). 

\end{document}